\documentclass[10pt,twocolumn,letterpaper]{article}

\usepackage{cvpr}
\usepackage{booktabs}
\usepackage{cuted}
\usepackage{caption}

  \usepackage{multirow}   
  \usepackage{xcolor}     
  \usepackage{amsmath}    
  \usepackage{amssymb}
\definecolor{cvprblue}{rgb}{0.21,0.49,0.74}
\usepackage[pagebackref,breaklinks,colorlinks,allcolors=cvprblue]{hyperref}

\def\confName{CVPR}
\def\confYear{2027}

\title{CarlHOI: Counterfactual Predictive Imagination for Generalizable Multi-Humanoid Object Interaction under Constrained Environments}

\author{
Wenjie Lin, Jin Wei Kocsis\thanks{Corresponding author.}\\
Purdue University\\
{\tt\small \{lin1790, kocsis0\}@purdue.edu}
}

\begin{document}
\maketitle

\begingroup
\setlength{\stripsep}{-9pt}
\begin{strip}

\centering
\includegraphics[width=0.92\textwidth]{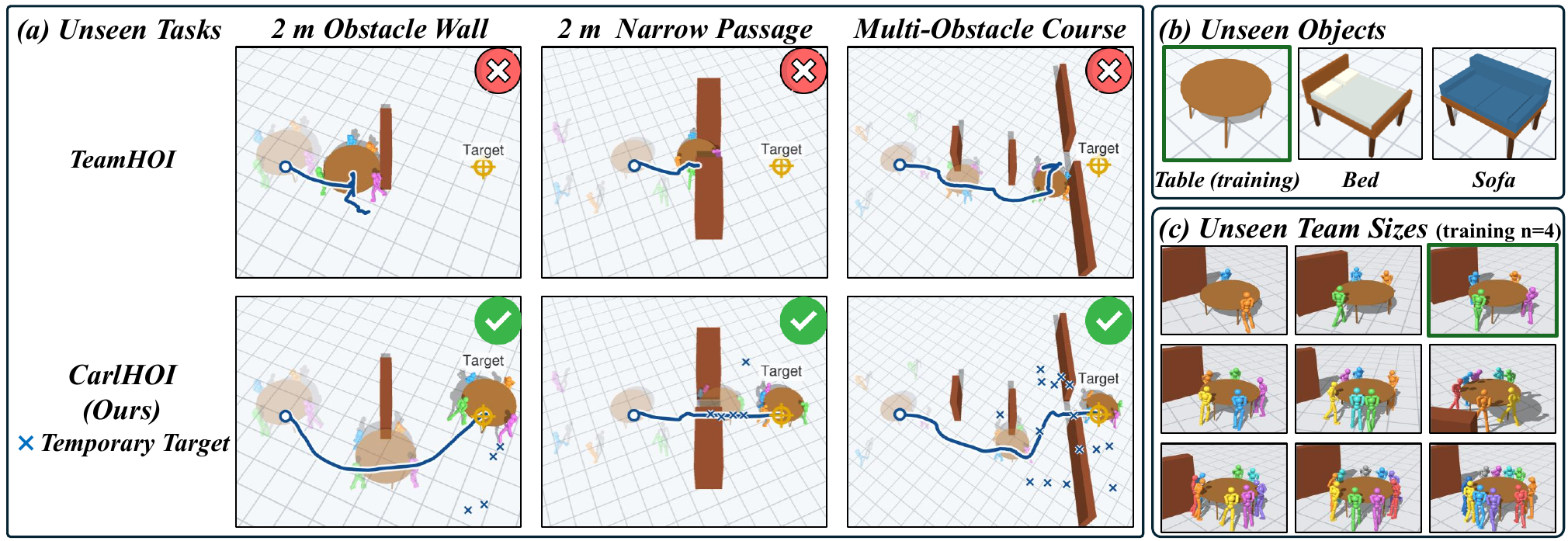}
\par
\captionsetup{
    hypcap=false,
    position=bottom,
}
\captionof{figure}{We present CarlHOI, a world model-based framework for multi-humanoid transport under challenges. We also introduce an obstacle benchmark spanning obstacle walls, narrow passages, and multi-obstacle courses with varied obstacle geometries and layouts, three objects, and nine team sizes. CarlHOI demonstrates superior transport performance and generalizes to different policies, unseen tasks, objects, and team sizes in zero-shot. Trajectories trace the table center, and blue crosses mark CarlHOI's temporary targets after pickup. Please refer to the \hyperref[sec:supp-results]{\textcolor{blue}{Supplementary Material}} for more intuitive visual results. More information is available on our \href{https://carlhoi.github.io/}{\textcolor{blue}{project website}}.} 
\label{fig:overview}
\vspace{16pt}
\end{strip}

\endgroup

\begin{figure*}[t]
    \centering
    \includegraphics[width=1\textwidth]{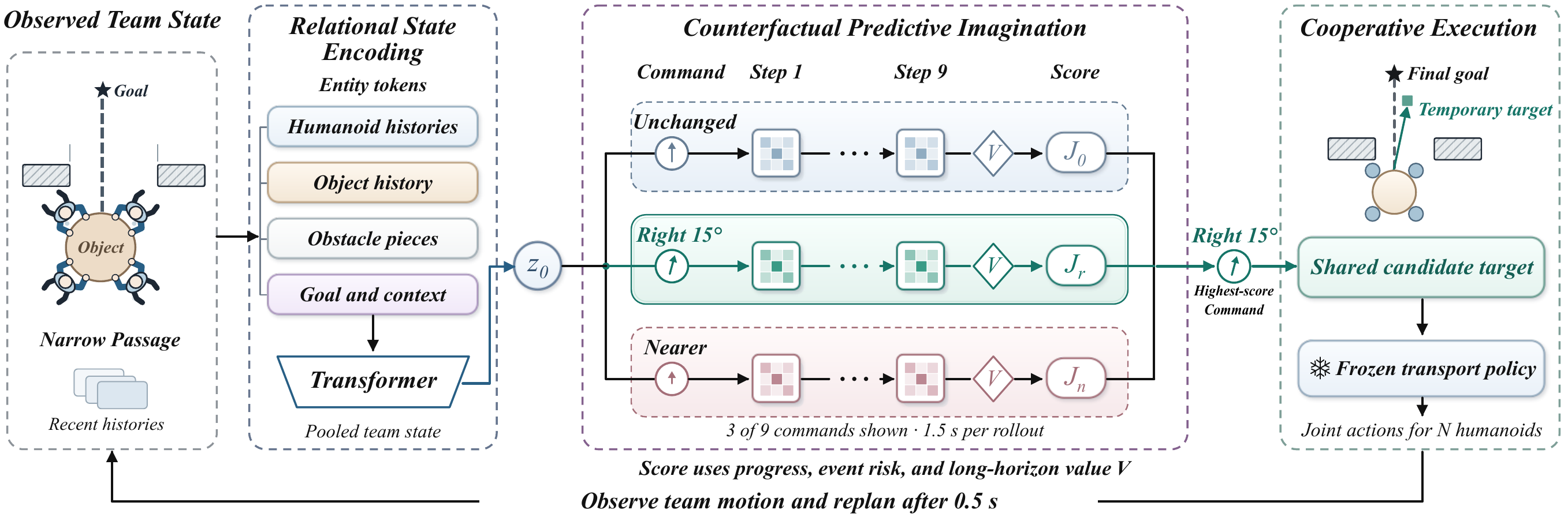}
    \caption{Overview of CarlHOI. CarlWM predicts how a humanoid team and its shared load will respond to alternative waypoint commands from the same observed state. Each prediction is anchored to guidance toward the original goal. The highlighted branch shows the waypoint selected for the frozen transport policy, which generates each humanoid's joint actions until CarlWM replans.}
    \label{fig:framework}
\end{figure*}
\begin{abstract}
Recent advances in physics-based humanoid control have enabled teams of varying sizes to coordinate their movements and cooperatively transport shared objects. However, adapting learned cooperative behaviors to unseen environmental constraints remains challenging, particularly when obstacles require coordinated detours and changes in team configuration. To address this challenge, we propose CarlHOI, a counterfactual predictive imagination framework built upon a counterfactual-anchored relational latent world model (CarlWM) for human-object interaction (HOI), enabling multi-humanoid teams to adapt to unseen environmental constraints without retraining their underlying cooperative control policies. CarlHOI explicitly represents the shared object, individual humanoids, and environmental obstacles as interacting entities. Trained offline using counterfactual simulator rollouts, the model imagines alternative collective futures under different shared waypoint commands, predicting how the coupled humanoid team and object would respond. By evaluating these imagined outcomes through predicted progress, failure, and contact risks, and long-horizon value estimates, CarlHOI selects adaptive guidance commands while preserving the pretrained cooperative controller. We introduce a benchmark comprising three challenging obstacle-constrained transport tasks to evaluate adaptation and generalization. Despite being trained exclusively with four humanoids, our world model generalizes without retraining to teams of two to ten humanoids, unseen obstacle geometries and placements, unseen objects, and novel combinations of environmental constraints. Extensive experiments demonstrate improved cooperative transport success over evaluated baselines across diverse configurations. These results highlight the potential of counterfactual predictive imagination to extend pretrained multi-humanoid cooperative skills beyond their original training conditions.

\end{abstract}

\section{Introduction}
\label{sec:intro}
Physics-based human-object interaction (HOI) requires coordinating body motion, contact, and object dynamics~\cite{wang2023physhoi,lu2025choice,jia2025primhoi,wang2025skillmimic}. In cooperative transport, the shared object further couples the humanoids' movements: each agent must respond to both the load and its teammates. Recent approaches have enabled such coordination across different objects and team sizes~\cite{gao2024coohoi,lionar2026teamhoi,lee2026mochi,deng2026ensemble}. However, constrained environments require the team to adapt both its route and its configuration around the object~\cite{lurz2022spline,patra2025kinodynamic,alvear2026cooperative}. Narrow passages may require humanoids to reposition, while blocking walls demand coordinated detours. These decisions must account for the entire team. 
For example, a collision-free path for the object may still cause a humanoid to collide with obstacles. We therefore study whether pretrained cooperative transport skills can meet these demands through shared temporary waypoints, avoiding repeated policy retraining for new layouts and team sizes. The central challenge is to predict how each waypoint affects the coupled team and object.

Hierarchical skill reuse~\cite{sharma2019dynamics,peng2019mcp,merel2020catch,peng2022ase,lin2025reflex,pandit2026decmht}
and predictive planning with world models~\cite{hansen2022tdmpc,hansen2024tdmpc2,zhao2026seqwm}
provide a basis for this adaptation. 
However, adapting cooperative transport to constrained environments remains challenging, as alternative waypoint commands can affect not only object motion but also the collective behavior of the humanoid team. Effective guidance must therefore consider object progress alongside team-level obstacle contacts and failure risks. To address this challenge, we introduce CarlHOI, a counterfactual predictive imagination framework built on a relational world model, CarlWM. CarlWM represents humanoids, the carried object, and obstacles as interacting entities and learns from counterfactual simulator rollouts to imagine how alternative waypoint commands affect the coupled team and object. Counterfactual anchoring expresses command effects relative to unchanged guidance. CarlHOI then selects shared waypoints using predicted progress, failure and contact risks, and long-horizon value estimates, while the frozen transport policy coordinates individual movements.

Our contributions are threefold:
\begin{itemize}
    \item \textbf{Counterfactual predictive guidance for cooperative transport.}
We propose CarlHOI and its relational world model, CarlWM, to predict and compare collective responses to alternative shared waypoints. Counterfactual anchoring, branched-rollout supervision, and long-horizon value estimation enable adaptive guidance of frozen transport policies.
    
    \item \textbf{A benchmark for constrained multi-humanoid transport.}
    We introduce narrow-passage, wall-detour, and multi-obstacle tasks that require coordinated transport under environmental constraints, with matched initial states and a common arrival criterion for systematic evaluation.

    \item \textbf{Zero-shot generalization to unseen teams, objects, and controllers.}
    Trained with four humanoids, one object, and one controller, CarlWM transfers without retraining to teams of 2-10, unseen objects, controllers, and obstacle configurations, improving transport success over evaluated baselines.
\end{itemize}
\section{Related Work}
\label{sec:related_work}

\paragraph{Physics-based HOI and cooperative transport.}
Motion imitation and adversarial motion priors underpin realistic humanoid control~\cite{peng2018deepmimic,peng2021amp,tessler2024maskedmimic,NEURIPS2024_215aeb07}. PhysHOI and InterMimic extend imitation to whole-body interactions~\cite{wang2023physhoi,xu2025intermimic}, while Catch \& Carry demonstrates reusable manipulation controllers~\cite{merel2020catch}. For cooperative transport, CooHOI learns coordination through shared object dynamics~\cite{gao2024coohoi}, while TeamHOI uses teammate tokens and a masked motion prior to support varying team sizes~\cite{lionar2026teamhoi}. The decMHT further addresses decentralized pickup and transport through object-centric control~\cite{pandit2026decmht}. In mobile manipulation, kinodynamic planners jointly account for robot motion, object transport, and obstacle constraints~\cite{patra2025kinodynamic}. However, adapting learned skills to environments that require coordinated
changes in both route and formation remains challenging without
retraining the transport policy. 
CarlHOI addresses this challenge by guiding pretrained transport policies through shared waypoints without retraining.

\paragraph{World models and relational prediction.}
TD-MPC and TD-MPC2 combine latent dynamics with learned terminal values for predictive control~\cite{hansen2022tdmpc,hansen2024tdmpc2}. SkiMo extends planning to skill-level dynamics~\cite{shi2022skill}, while Freed et al.\ learn skills and temporally abstract world models from offline data~\cite{freed2023abstract}. For multi-agent cooperation, SeqWM uses sequential agent-wise predictions~\cite{zhao2026seqwm}, and COMBO predicts joint action outcomes with compositional generative models~\cite{zhang2025combo}. Relational models represent physical systems through entities and their interactions, supporting structured dynamics prediction~\cite{battaglia2016interaction,kipf2018neural,li2018learning,sanchezgonzalez2020gns,kipf2019contrastive}. 
For cooperative transport, a key challenge is to evaluate how alternative shared waypoints affect the coupled motion of humanoids and their carried object. CarlWM addresses this challenge through entity-based encoding that captures their collective response to waypoint commands. It learns from counterfactual rollouts starting at matched simulator states and anchors command effects to unchanged guidance, enabling predictions that support the comparison of candidate waypoints. 
\section{Problem Formulation}
\label{sec:problem}

A team of $n$ humanoids must lift an object and carry it to a goal $g\in\mathbb{R}^2$ 
through an obstacle-constrained environment. An episode succeeds if the object's center $p_t$ comes within a tolerance of $g$ before the time limit, without a
humanoid falling or the object being dropped.

Each humanoid executes a shared pretrained transport policy $\pi$, conditioned on a shared goal, to generate its joint actions. We keep $\pi$ frozen and influence the team only through the goal it observes. At decision times
$k=0,1,\dots$, a planner replaces $g$ with a temporary aim point
\begin{equation}
  \hat g_k=p_t+\sigma_k R(\theta_k)\,(g-p_t),\qquad (\theta_k,\sigma_k)\in\mathcal{A},
  \label{eq:aim}
\end{equation}
where $\theta_k$ and $\sigma_k$ specify the rotation and rescaling of the goal direction, respectively. The selected command is held fixed until the next decision. The planner thus decides \emph{where} the team goes, and the frozen
policy decides \emph{how} it carries the object there.

The planner's objective is to maximize the expected return of the per-step reward
\begin{equation}
  r_t=(d_{t-1}-d_t)-w_f\,f_t-w_c\,c_t,
  \label{eq:reward}
\end{equation}
where $d_t=\lVert p_t-g\rVert$ is the distance to the true goal, $f_t$ and $c_t$ indicate a failure and a wall contact,
and $w_f,w_c>0$. 
The predictive model is trained under limited conditions and must generalize without retraining to unseen team sizes, obstacles, objects, and transport policies.

\providecommand{\ours}{CarlHOI}
\providecommand{\teamhoi}{TeamHOI}
\providecommand{\teamhoift}{TeamHOI-FT}
\providecommand{\coohoi}{CooHOI}
\providecommand{\tdmpc}{TD-MPC2}
\providecommand{\seqwm}{SeqWM}

\providecommand{\carlwm}{CarlWM}

\section{The CarlHOI Framework}
\label{sec:methodology}

\ours{} instantiates the planner of \cref{sec:problem} with a discrete command library and a learned world model.

\paragraph{Command library.}
CarlHOI restricts the admissible set $\mathcal{A}=[-45^\circ,45^\circ]\times[0.5,1.5]$ to a library $\mathcal{U}$ of
nine commands $u=(\theta,\sigma)$: the unchanged goal $u^{0}=(0^\circ,1)$, rotations of $\pm15^\circ$, $\pm30^\circ$ and
$\pm45^\circ$, and rescalings by $0.5$ and $1.5$. CarlHOI selects a command every $\Delta=0.5$\,s; the aim
distance~\eqref{eq:aim} of every command except $u^{0}$ is clipped to 0.5--9\,m, while $u^{0}$ passes the true goal
unchanged. 
\carlwm{} is activated after object pickup; before that, the policy sees the true goal. 
The model operates at $1/6$\,s intervals. We write $r_\tau$ for the reward~\eqref{eq:reward} accumulated over model step $\tau$, with $w_f=3$ and $w_c=0.3/H$. A command
is executed for three model steps but imagined over a horizon of $H=9$ steps (1.5\,s).

\paragraph{Overview.}
As illustrated in Fig. 2, at each decision, \carlwm{} encodes recent scene history into a latent state (\cref{sec:encoder}), imagines
alternative command outcomes over $H$ steps (\cref{sec:dynamics}), and estimates its return using the predicted rewards and a
learned long-horizon value (\cref{sec:planning}). 
CarlHOI executes the command with the highest estimated return and replans. \carlwm{} is trained offline on counterfactual simulator rollouts (\cref{sec:training}). The encoder
represents humanoids and obstacle pieces as tokens with shared weights, so the same parameters can process any number
of them. Whether the model transfers across team sizes, obstacles, objects and policies, as required in
\cref{sec:problem}, is established empirically (\cref{sec:Results}).

\subsection{Relational entity encoder}
\label{sec:encoder}

\paragraph{Observation.}
At time $t$, the scene consists of the object state $x_t$ (position $p_t$, height, heading, velocity, and up-vector),
the state $\xi^i_t$ of each humanoid $i$ (position relative to the object, height, heading, velocity, and the distance of
each hand to the nearest grip point), and the obstacle pieces $\mathcal{B}=\{b_j\}_{j=1}^{m}$, each an oriented box;
we write $s_t=(x_t,\xi^{1:n}_t,\mathcal{B})$. 
At decision $k$, the planner observes $o_k=\big(s_{\tau-L+1:\tau},\,\mathbf{a}_{\tau-L+1:\tau-1},\,g,\,\tau\big)$, comprising the last $L=6$ scene states (spanning $5/6$\,s), intervening commands, the goal, and elapsed time, where $\tau$ is the corresponding model step. A command 
is encoded as
$\mathbf{a}=(\theta/15^\circ,\ \log_2\sigma,\ \alpha/8)$, where $\alpha$ denotes its age in model steps, distinguishing newly issued commands from ongoing guidance.

\paragraph{Entity tokens.}
We encode $o_k$ into entity and context tokens using type-specific linear maps:
\begin{align}
  e^{\mathrm{obj}}&=W_{\mathrm{obj}}\big[x_{\tau-L+1:\tau},\,g-p_\tau,\,\lVert g-p_\tau\rVert\big]+\eta_{\mathrm{obj}},\\
  e^{i}_{\mathrm{hum}}&=W_{\mathrm{hum}}\big[\xi^{i}_{\tau-L+1:\tau},\,\cos n\phi^{i},\,\sin n\phi^{i}\big]+\eta_{\mathrm{hum}},\\
  e^{j}_{\mathrm{obs}}&=W_{\mathrm{obs}}\big[\beta^{j}_{\tau-L+1:\tau},\,\mathrm{size}(b_j)\big]+\eta_{\mathrm{obs}},\\
  e^{\mathrm{ctx}}&=W_{\mathrm{ctx}}\big[\mathbf{a}_{\tau-L+1:\tau-1},\,\tau\big]+\eta_{\mathrm{ctx}},
\end{align}
for $i=1,\dots,n$ and $j=1,\dots,m$,
where $\eta$ are learned type embeddings. Humanoid quantities are expressed relative to the object. $\phi^{i}$ is the
bearing of humanoid $i$ around the object, and the $n$-fold features $(\cos n\phi^{i},\sin n\phi^{i})$ encode its angular phase
relative to an $n$-fold symmetric arrangement, so that all $n$ symmetric carrying positions share the same value. 
Each obstacle piece is represented by its object-relative box history $\beta^{j}$, comprising near and far faces along the route and the two lateral edges, and by its size $\mathrm{size}(b_j)$. Each lateral offset $d_{\mathrm{lat}}$ is saturated as $r_{\mathrm{sat}}\tanh(d_{\mathrm{lat}}/r_{\mathrm{sat}})$, with $r_{\mathrm{sat}}=3$\,m, reducing sensitivity to distant obstacles.

\paragraph{Aggregation.}
The tokens, together with a learned pooling token $e^{\mathrm{pool}}$, are processed by a transformer encoder $T$
(3 pre-norm layers, 4 heads, width 128) without positional encodings, and absent humanoid and obstacle slots are
masked out. The output of the pooling token is mapped to the latent state,
\begin{equation}
  \begin{aligned}
    \mathbf{t}&=T\big(e^{\mathrm{obj}},e^{1:n}_{\mathrm{hum}},e^{1:m}_{\mathrm{obs}},e^{\mathrm{ctx}},e^{\mathrm{pool}}\big),\\
    z_k&=\mathrm{SimNorm}\big(\mathrm{MLP}(\mathbf{t}_{\mathrm{pool}})\big),
  \end{aligned}
  \label{eq:encoder}
\end{equation}
with $z_k\in\mathbb{R}^{128}$,
where $\mathrm{SimNorm}$ projects groups of latent dimensions onto simplices
~\cite{hansen2024tdmpc2}. Since 
humanoid tokens share $W_{\mathrm{hum}}$ and 
carry no indices, $z_k$ is invariant to permutations of the
humanoids, and the same parameters can process varying team sizes and obstacle counts. 
Trained with four humanoid slots, the model accommodates up to ten at inference without retraining; absent humanoid slots are masked (details in Supplementary Material,
Sec.~\ref{sec:supp-carlwm}).

\subsection{Latent dynamics and prediction heads}
\label{sec:dynamics}

From $z$, \carlwm{} predicts the next latent state, 
progress, and event probabilities under command encoding $\mathbf{a}$:
\begin{gather}
  z'=\mathrm{SimNorm}\big(D(z)+C(z,\mathbf{a})-C(z,\mathbf{a}^{0})\big),
  \label{eq:dynamics}\\
  (\hat\delta,\ell^{f},\ell^{c},\ell^{a})
  =Y(z)+E(z,\mathbf{a})-E(z,\mathbf{a}^{0}),
  \label{eq:heads}
\end{gather}
where $\hat\delta$ is the predicted progress $\delta_\tau=d_{\tau-1}-d_\tau$, $\ell^{f},\ell^{c},\ell^{a}$ are the logits of failure, wall contact and
arrival with probabilities $\hat q^{\{f,c,a\}}=\mathrm{sigmoid}(\ell^{\{f,c,a\}})$, and $D,C,Y,E$ are MLPs.
$\mathbf{a}^{0}$ is the unchanged command with the same command age. This \emph{counterfactual anchoring} expresses the predicted effect
of a command as a deviation from the policy's default response, which is learned from all data, and makes the
prediction for $u^{0}$ independent of $C$ and $E$. Rolling~\eqref{eq:dynamics} forward from $z_k$ with 
command encodings $\mathbf{a}_{1:H}$ over $H$ steps gives an imagined trajectory
$z_{k}=\hat z_0,\hat z_1,\dots,\hat z_H$ with per-step predictions $(\hat\delta_h,\hat q_h^{f},\hat q_h^{c},\hat q_h^{a})$.

\subsection{Long-horizon value and planning}
\label{sec:planning}

A horizon of $1.5$\,s 
may not capture the benefit of a detour around a wall, which first increases the distance to
the goal. \carlwm{} therefore estimates the expected discounted return of~\eqref{eq:reward}, the planner's objective
(\cref{sec:problem}), 
using the predicted rewards over the horizon and an ensemble of $N_V=5$ value heads $V_l(z)$ to estimate returns beyond it. A command $u$ is scored by
\begin{equation}
  \begin{aligned}
    J(u)={}&\sum_{h=1}^{H}\gamma^{h-1}A_{h-1}\big(\hat\delta_h-w_f\hat q^{f}_h-w_c\hat q^{c}_h\big)\\
    &+\lambda\gamma^{H}A_{H}\,\bar V(\hat z_H)+\omega\,\mathbf{1}[u=u^{0}].
  \end{aligned}
  \label{eq:score}
\end{equation}
Here, $\hat q_h^{\mathrm{end}}
=1-(1-\hat q_h^{f})(1-\hat q_h^{a})$, and
$A_h=\prod_{h'=1}^{h}(1-\hat q_{h'}^{\mathrm{end}})$
is an estimated continuation weight, with $A_0=1$.
We use $\bar V=\frac{1}{N_V}\sum_{l=1}^{N_V}V_l$,
$\gamma=0.97$, and $\lambda=1$.
The reference bonus $\omega=0.02$, equivalent to 2\,cm under
the progress reward scale, favors unchanged guidance
when predicted returns are close.
The first term estimates the discounted return over the
horizon, and the second adds the value of the imagined
end state, weighted by $A_H$.
The executed command is
\begin{equation}
  u_k^\star=\arg\max_{u\in\mathcal{U}}J(u),
  \label{eq:argmax}
\end{equation}
computed by evaluating all $|\mathcal{U}|=9$ imagined trajectories in one batch.

\subsection{Counterfactual data and training}
\label{sec:training}

\paragraph{Branched rollouts.}
A world model that only observes the executed behavior rarely sees the consequences of the commands it will compare.
We therefore collect \emph{counterfactual} data with the simulator. Along episodes in which the policy is steered by a
mixture of commands, we snapshot the full simulator state at regular intervals and simulate several commands of the library for $H$
steps from that snapshot (five or all nine, depending on the collection round). Every branch point thus provides the outcomes of several commands from an identical state.

\paragraph{Objective.}
For a sample with history $o_k$, command encodings $\mathbf{a}_{1:H}$, observed per-step targets
$(\delta_h,f_h,c_h,\mathrm{arr}_h)$, and future observations $o_{k,h}$, we roll out~\eqref{eq:dynamics} from $z_k$ and
minimize
\begin{equation}
  \mathcal{L}=w_{z}\mathcal{L}_{z}+w_{\delta}\mathcal{L}_{\delta}+w_{q}\mathcal{L}_{q}+w_{V}\mathcal{L}_{V}+w_{\mathrm{rk}}\mathcal{L}_{\mathrm{rank}},
  \label{eq:loss}
\end{equation}
with $(w_z,w_\delta,w_q,w_V,w_{\mathrm{rk}})=(20,1,1,0.5,1)$ and the terms
\begin{align}
  \mathcal{L}_{z}&=\big\langle\big\lVert\hat z_h-\mathrm{sg}\big(\mathrm{enc}(o_{k,h})\big)\big\rVert_2^2\big\rangle_{\rho,\bar\mu},\\
  \mathcal{L}_{\delta}&=\big\langle\big((\hat\delta_h-\delta_h)/\nu\big)^2\big\rangle_{\rho,\mu},\\
  \mathcal{L}_{q}&=\big\langle\tfrac13\textstyle\sum_{\kappa\in\{f,c,a\}}\mathrm{BCE}\big(\ell^{\kappa}_h,\kappa_h\big)\big\rangle_{\rho,\mu},\\
  \mathcal{L}_{V}&=\textstyle\frac{1}{N_V}\sum_{l=1}^{N_V}\big\langle\big(V_l(\hat z_h)-G_{k,h}\big)^2\big\rangle_{1,\mu^{V}},
\end{align}
where
$\langle \psi\rangle_{\rho,\mu}=\sum_{k,h}\rho^{h-1}\mu_{k,h}\psi_{k,h}\big/\sum_{k,h}\rho^{h-1}\mu_{k,h}$
is a weighted average over the samples $k$ of a minibatch and the steps $h=1,\dots,H$ ($h=0,\dots,H$ for
$\mathcal{L}_V$, without temporal weighting), $\rho=0.9$ down-weights distant steps, $\mu$ masks steps after the
episode ends, $\bar\mu$ also masks the terminal step, $\nu=0.05$\,m expresses progress in units of 5\,cm, $\mathrm{sg}$
stops gradients, and $\mathrm{enc}$ is the encoder of~\eqref{eq:encoder}.
The value targets come from the executed episodes:
$G_{k,h}=\sum_{\tau'\ge1}\gamma^{\tau'-1}r_{\tau+h+\tau'}$ is the discounted return of the reward~\eqref{eq:reward}
accrued after step $\tau+h$, summed to the end of the recorded episode without bootstrapping. The mask $\mu^{V}$ selects
the non-terminal states with such a target: steps $h=0,\dots,H$ of samples along executed trajectories, rolled out using
the recorded commands so that predictions and return targets correspond to the same command history and time step, and
only the branch point ($h=0$) of branched samples, so no return beyond an $H$-step branch is needed. 
The value heads thus learn the returns achieved by the frozen policy under recorded commands.

\paragraph{Ranking loss.}
The regression terms supervise individual rollouts, but the planner relies on comparisons among commands; the ranking loss supervises these comparisons directly. At
each branch point $\mathcal{P}$ with simulated commands $\mathcal{U}_{\mathcal{P}}$, we compute for every
$u\in\mathcal{U}_{\mathcal{P}}$ the model's survival-weighted horizon score
$J^{\circ}(u)=\sum_{h=1}^{H}A_{h-1}\big(\hat\delta_h-w_f\hat q^{f}_h-w_c\hat q^{c}_h\big)$ and the realized horizon return
$G^{\circ}(u)=\sum_{h=1}^{H}r_h$ of the simulated branch, and regress the centered scores on the centered returns:
\begin{equation}
\resizebox{\columnwidth}{!}{$\displaystyle
\mathcal{L}_{\mathrm{rank}}
=
\frac{1}{|\mathcal{M}|}
\sum_{\mathcal{P}\in\mathcal{M}}
\frac{1}{|\mathcal{U}_{\mathcal{P}}|}
\sum_{u\in\mathcal{U}_{\mathcal{P}}}
\mathrm{Huber}_{\epsilon}\!\left(
\tfrac{J^{\circ}(u)-\overline{J^{\circ}}_{\mathcal{P}}}{\nu},
\tfrac{G^{\circ}(u)-\overline{G^{\circ}}_{\mathcal{P}}}{\nu}
\right)
$},
\label{eq:rank}
\end{equation}
where $\mathcal{M}$ is a minibatch of branch points, $\overline{(\cdot)}_{\mathcal{P}}$ is the mean over
$\mathcal{U}_{\mathcal{P}}$, $\nu$ expresses the centered horizon scores in units of 5\,cm of progress, and
$\mathrm{Huber}_{\epsilon}$ is quadratic for errors below $\epsilon=4$ and linear above, which limits the influence of rare
failure branches. Centering removes the state-dependent offset, which does not affect the choice, but keeps the gaps
between commands, which matter once the value and the bonus in~\eqref{eq:score} are added. Since $J^{\circ}$ and $G^{\circ}$
exclude discounting, the value and the bonus, the loss supervises the short-horizon comparison of commands that the
branched data observe directly; the long-horizon part of $J$ is learned by $\mathcal{L}_V$. The loss requires several
commands simulated from the same state, which only the branched data provide.


\providecommand{\ours}{CarlHOI}
\providecommand{\teamhoi}{TeamHOI}
\providecommand{\teamhoift}{TeamHOI-FT}
\providecommand{\coohoi}{CooHOI$^\ast$}
\providecommand{\coohoift}{CooHOI$^\ast$-FT}
\providecommand{\tdmpc}{TD-MPC2}
\providecommand{\seqwm}{SeqWM}
\providecommand{\carlwm}{CarlWM}
\providecommand{\teamhoi}{TeamHOI}
\providecommand{\teamhoift}{TeamHOI-FT}
\providecommand{\tdmpc}{TD-MPC2}
\providecommand{\seqwm}{SeqWM}
\providecommand{\coohoi}{CooHOI}

\begin{table*}[t]
\centering
\caption{\textbf{Carrying the table across policies, tasks, and team sizes.} Each team size $n$ is split into W: Obstacle Wall (spans 1.0--3.0\,m), P: Narrow Passage (2.1 and 2.0\,m gates) and M: Multi-Obstacle Course (two 13\,m layouts).  Each obstacle is evaluated on 256 episodes per team size. \textbf{$^\dagger$Planners are only trained with $n=4$ and \teamhoift{} on 2.1\,m passages and 1.25-1.75\,m walls; every other policy, team size, and obstacle task is zero-shot.} Time to arrival: mean arrival time over successful episodes of the task; --: no successful episode. \textbf{Detailed results for individual subtasks, objects, policies,
and team sizes are provided in the Supplementary Material,
Sec.~\ref{sec:supp-results}.}}
\label{tab:main-table-object}
\setlength{\tabcolsep}{1.3pt}
\resizebox{\textwidth}{!}{%
\begin{tabular}{@{}ll@{\hspace{5pt}}ccc@{\hspace{5pt}}ccc@{\hspace{5pt}}ccc@{\hspace{5pt}}ccc@{\hspace{5pt}}ccc@{\hspace{5pt}}ccc@{\hspace{5pt}}ccc@{\hspace{5pt}}ccc@{\hspace{5pt}}ccc@{\hspace{5pt}}ccc@{}}
\toprule
 & & \multicolumn{27}{c}{Team size $n$} & \\
\cmidrule(lr){3-29}
 & & \multicolumn{3}{c}{2} & \multicolumn{3}{c}{3} & \multicolumn{3}{c}{4$^\dagger$} & \multicolumn{3}{c}{5} & \multicolumn{3}{c}{6} & \multicolumn{3}{c}{7} & \multicolumn{3}{c}{8} & \multicolumn{3}{c}{9} & \multicolumn{3}{c}{10} & \multicolumn{3}{c}{Mean} \\
\cmidrule(lr){3-5}\cmidrule(lr){6-8}\cmidrule(lr){9-11}\cmidrule(lr){12-14}\cmidrule(lr){15-17}\cmidrule(lr){18-20}\cmidrule(lr){21-23}\cmidrule(lr){24-26}\cmidrule(lr){27-29}\cmidrule(lr){30-32}
Policy & Planner & W & P & M & W & P & M & W & P & M & W & P & M & W & P & M & W & P & M & W & P & M & W & P & M & W & P & M & W & P & M \\
\midrule
\multicolumn{32}{@{}l}{\textit{\textbf{Success rate (\%) $\uparrow$}}} \\
\midrule
\multirow{4}{*}{\teamhoi{}} & Policy alone & 51.1 & 54.5 & 21.5 & 31.3 & 31.8 & 12.1 & 42.7 & 39.0 & 14.3 & 48.1 & 34.8 & 14.8 & 43.8 & 27.0 & 14.8 & 44.6 & 22.5 & 13.3 & 45.3 & \textbf{23.0} & 18.8 & 45.1 & 19.3 & 14.1 & 47.2 & 17.2 & 14.1 & 44.4 & 29.9 & 15.3 \\
 & \tdmpc{} & 14.2 & 11.1 & 0.8 & 24.6 & 5.9 & 1.2 & 32.9 & 9.8 & 2.5 & 31.4 & 7.4 & 2.3 & 30.1 & 5.1 & 2.0 & 28.8 & 3.7 & 1.2 & 29.1 & 4.1 & 2.0 & 28.3 & 3.5 & 0.8 & 25.9 & 2.5 & 0.4 & 27.3 & 5.9 & 1.5 \\
 & \seqwm{} & 30.1 & 7.2 & 24.2 & 29.6 & 2.1 & 7.4 & 29.7 & 23.7 & 4.7 & 27.6 & 24.4 & 6.6 & 31.2 & 18.2 & 6.2 & 28.8 & 16.6 & 6.2 & 25.4 & 11.7 & 3.5 & 24.9 & 9.8 & 3.1 & 25.7 & 8.0 & 3.9 & 28.1 & 13.5 & 7.3 \\
 & \textbf{\carlwm{} (Ours)} & \textbf{84.7} & \textbf{61.7} & \textbf{39.8} & \textbf{80.5} & \textbf{32.6} & \textbf{29.7} & \textbf{81.2} & \textbf{47.9} & \textbf{32.4} & \textbf{78.7} & \textbf{42.4} & \textbf{28.9} & \textbf{75.6} & \textbf{30.1} & \textbf{24.6} & \textbf{77.5} & \textbf{24.8} & \textbf{21.9} & \textbf{78.1} & 22.7 & \textbf{20.3} & \textbf{82.0} & \textbf{21.1} & \textbf{26.6} & \textbf{84.4} & \textbf{19.5} & \textbf{26.2} & \textbf{80.3} & \textbf{33.6} & \textbf{27.8} \\
\cmidrule(l){2-32}
\multirow{4}{*}{\teamhoift{}$^\dagger$} & Policy alone & 25.2 & 68.9 & 41.0 & 38.7 & 41.8 & 23.0 & 49.7 & 55.4 & 32.2 & 52.7 & 51.8 & 31.2 & 55.6 & 45.7 & 24.6 & 58.4 & 40.8 & 19.5 & 58.3 & 34.0 & 9.4 & 57.9 & 35.7 & 10.2 & 59.2 & 27.1 & 7.0 & 50.6 & 44.6 & 22.0 \\
 & \tdmpc{} & 12.6 & 7.2 & 0.8 & 24.7 & 5.1 & 3.5 & 37.5 & 12.2 & 2.0 & 38.8 & 12.1 & 2.0 & 32.9 & 9.8 & 0.4 & 33.3 & 8.0 & 1.6 & 31.4 & 6.8 & 0.4 & 28.6 & 4.5 & 0.0 & 26.5 & 4.3 & 0.4 & 29.6 & 7.8 & 1.2 \\
 & \seqwm{} & 28.4 & 7.6 & 19.5 & 32.0 & 3.7 & 14.8 & 32.7 & 31.3 & 10.9 & 35.0 & 27.7 & 5.9 & 34.5 & 27.7 & 10.2 & 30.7 & 24.8 & 10.5 & 29.9 & 20.9 & 5.5 & 27.8 & 19.1 & 3.9 & 26.7 & 16.2 & 4.3 & 30.9 & 19.9 & 9.5 \\
 & \textbf{\carlwm{} (Ours)} & \textbf{86.4} & \textbf{74.6} & \textbf{46.1} & \textbf{85.4} & \textbf{49.8} & \textbf{42.2} & \textbf{91.6} & \textbf{68.0} & \textbf{52.5} & \textbf{90.1} & \textbf{63.1} & \textbf{48.8} & \textbf{89.7} & \textbf{54.9} & \textbf{52.0} & \textbf{90.3} & \textbf{50.8} & \textbf{50.0} & \textbf{89.5} & \textbf{49.8} & \textbf{43.8} & \textbf{89.9} & \textbf{47.3} & \textbf{48.8} & \textbf{88.0} & \textbf{41.2} & \textbf{43.4} & \textbf{89.0} & \textbf{55.4} & \textbf{47.5} \\
\cmidrule(l){2-32}
\multirow{4}{*}{\coohoi{}} & Policy alone & 53.3 & 35.9 & 7.4 & 63.6 & 7.0 & 10.2 & 68.2 & 9.2 & 2.5 & 55.0 & 20.5 & 13.7 & 42.6 & 9.2 & 16.8 & 37.3 & 6.1 & 1.2 & 26.2 & 13.9 & 0.4 & 21.9 & 7.8 & 0.0 & 16.9 & 4.5 & 0.0 & 42.8 & 12.7 & 5.8 \\
 & \tdmpc{} & 19.2 & 8.0 & 1.2 & 24.0 & 1.0 & 0.4 & 31.0 & 5.7 & 0.6 & 29.3 & 2.0 & 1.2 & 22.2 & 1.4 & 0.4 & 18.5 & 1.0 & 0.0 & 15.0 & 0.6 & 0.0 & 12.8 & 0.2 & 0.0 & 11.9 & 0.4 & 0.0 & 20.4 & 2.2 & 0.4 \\
 & \seqwm{} & 21.3 & 6.8 & 10.5 & 23.2 & 0.2 & 2.3 & 40.8 & 10.3 & 2.7 & 27.4 & 10.7 & 6.2 & 21.0 & 4.1 & 3.9 & 14.4 & 2.7 & 0.4 & 14.2 & 6.8 & 2.3 & 9.5 & 1.4 & 0.8 & 6.2 & 1.6 & 0.0 & 19.8 & 5.0 & 3.3 \\
 & \textbf{\carlwm{} (Ours)} & \textbf{86.1} & \textbf{60.5} & \textbf{35.5} & \textbf{83.0} & \textbf{12.5} & \textbf{10.9} & \textbf{83.7} & \textbf{25.8} & \textbf{18.6} & \textbf{81.4} & \textbf{32.6} & \textbf{32.4} & \textbf{77.4} & \textbf{18.8} & \textbf{27.7} & \textbf{75.6} & \textbf{12.7} & \textbf{5.9} & \textbf{74.7} & \textbf{14.3} & \textbf{10.5} & \textbf{63.0} & \textbf{13.1} & \textbf{4.3} & \textbf{61.5} & \textbf{9.8} & 0.0 & \textbf{76.3} & \textbf{22.2} & \textbf{16.2} \\
\midrule
\multicolumn{32}{@{}l}{\textit{\textbf{Time to arrival (s), successful episodes only $\downarrow$} }} \\
\midrule
\multirow{4}{*}{\teamhoi{}} & Policy alone & 22.6 & \textbf{14.3} & 41.7 & 21.3 & 18.0 & 40.6 & 21.0 & 15.8 & 39.7 & 20.7 & 16.7 & 41.9 & 20.8 & 17.6 & 39.2 & 20.8 & \textbf{17.9} & 41.5 & 20.8 & 19.1 & 41.8 & 20.9 & 19.4 & 41.3 & 20.8 & 20.0 & 41.1 & 21.1 & 17.7 & 41.0 \\
 & \tdmpc{} & 22.6 & 23.6 & \textbf{27.3} & 22.7 & 24.6 & 41.8 & 22.1 & 23.3 & 44.3 & 21.8 & 23.7 & 43.0 & 22.3 & 25.1 & 46.3 & 22.7 & 24.6 & 41.8 & 22.6 & 26.1 & 45.5 & 22.9 & 25.7 & 45.1 & 23.5 & 24.5 & 37.7 & 22.6 & 24.6 & 41.4 \\
 & \seqwm{} & 23.0 & 25.1 & 41.9 & 23.6 & 23.8 & \textbf{35.8} & 24.5 & 22.6 & 38.8 & 23.7 & 23.9 & 37.7 & 24.2 & 23.2 & 38.5 & 24.1 & 22.8 & 40.9 & 24.7 & 24.4 & 40.7 & 24.5 & 24.7 & 40.5 & 25.0 & 24.2 & 42.2 & 24.2 & 23.9 & 39.7 \\
 & \textbf{\carlwm{} (Ours)} & \textbf{15.6} & \textbf{14.3} & 28.5 & \textbf{15.1} & \textbf{17.3} & 36.5 & \textbf{14.7} & \textbf{14.6} & \textbf{34.0} & \textbf{14.6} & \textbf{15.5} & \textbf{32.9} & \textbf{14.9} & \textbf{16.7} & \textbf{33.1} & \textbf{15.5} & 18.0 & \textbf{35.3} & \textbf{16.0} & \textbf{18.7} & \textbf{38.4} & \textbf{16.4} & \textbf{18.8} & \textbf{37.1} & \textbf{16.5} & \textbf{18.2} & \textbf{37.2} & \textbf{15.5} & \textbf{16.9} & \textbf{34.8} \\
\cmidrule(l){2-32}
\multirow{4}{*}{\teamhoift{}$^\dagger$} & Policy alone & 23.5 & 17.6 & 40.5 & 21.4 & 19.0 & 41.3 & 21.5 & 16.3 & 40.2 & 21.0 & 17.0 & 41.1 & 21.2 & 18.0 & 43.0 & 21.5 & 18.9 & 42.9 & 21.5 & 20.2 & 42.6 & 21.5 & 20.2 & 45.8 & 21.6 & 20.0 & 46.3 & 21.6 & 18.6 & 42.6 \\
 & \tdmpc{} & 24.0 & 24.9 & 47.9 & 23.3 & 25.2 & 42.5 & 23.0 & 24.6 & 45.0 & 22.9 & 24.9 & 43.1 & 23.1 & 24.8 & 35.0 & 23.3 & 25.4 & 48.6 & 23.5 & 25.4 & 49.0 & 23.5 & 25.7 & -- & 24.4 & 25.4 & 39.7 & 23.4 & 25.2 & 43.9 \\
 & \seqwm{} & 24.4 & 25.8 & 42.8 & 24.0 & 26.9 & 36.9 & 24.7 & 23.3 & 40.1 & 24.2 & 23.6 & 37.2 & 24.6 & 23.7 & 38.6 & 25.0 & 24.2 & 40.4 & 25.3 & 24.4 & 41.3 & 25.5 & 24.4 & 42.1 & 25.6 & 24.6 & 43.5 & 24.8 & 24.5 & 40.3 \\
 & \textbf{\carlwm{} (Ours)} & \textbf{17.3} & \textbf{16.4} & \textbf{32.9} & \textbf{16.3} & \textbf{18.6} & \textbf{36.8} & \textbf{15.5} & \textbf{14.6} & \textbf{34.6} & \textbf{15.4} & \textbf{16.2} & \textbf{35.3} & \textbf{15.6} & \textbf{16.9} & \textbf{34.0} & \textbf{16.1} & \textbf{18.2} & \textbf{36.3} & \textbf{16.7} & \textbf{18.3} & \textbf{36.9} & \textbf{17.1} & \textbf{18.9} & \textbf{38.2} & \textbf{17.2} & \textbf{19.3} & \textbf{38.0} & \textbf{16.4} & \textbf{17.5} & \textbf{35.9} \\
\cmidrule(l){2-32}
\multirow{4}{*}{\coohoi{}} & Policy alone & 21.1 & 18.8 & 43.9 & 20.6 & \textbf{24.5} & 45.6 & 21.1 & 19.3 & 43.4 & 21.0 & 21.9 & 42.7 & 22.0 & 23.5 & 44.0 & 22.0 & 23.5 & 50.2 & 23.1 & 24.7 & \textbf{41.4} & 22.8 & 24.1 & -- & 23.8 & 24.2 & -- & 21.9 & 22.7 & 44.5 \\
 & \tdmpc{} & 22.6 & 23.8 & \textbf{32.6} & 24.0 & 26.7 & 46.2 & 23.0 & 23.7 & 47.2 & 23.3 & 26.6 & 46.5 & 23.2 & 26.1 & 44.3 & 23.0 & 27.0 & -- & 24.3 & 24.7 & -- & 23.3 & 28.0 & -- & 24.2 & 27.5 & -- & 23.4 & 26.0 & 43.3 \\
 & \seqwm{} & 24.4 & 26.1 & 43.6 & 25.0 & 29.3 & \textbf{41.6} & 23.8 & 21.8 & 41.2 & 23.8 & 24.3 & 39.7 & 24.4 & 23.7 & 43.1 & 25.4 & 24.2 & \textbf{42.4} & 25.4 & 24.6 & 43.5 & 25.2 & 24.3 & \textbf{44.4} & 25.5 & 25.7 & -- & 24.8 & 24.9 & 42.4 \\
 & \textbf{\carlwm{} (Ours)} & \textbf{19.3} & \textbf{16.5} & 36.8 & \textbf{20.0} & 24.9 & 43.8 & \textbf{16.9} & \textbf{16.8} & \textbf{36.7} & \textbf{17.9} & \textbf{21.2} & \textbf{38.7} & \textbf{18.9} & \textbf{21.9} & \textbf{43.0} & \textbf{19.8} & \textbf{22.9} & 45.1 & \textbf{20.7} & \textbf{21.4} & 45.7 & \textbf{21.0} & \textbf{21.8} & 46.3 & \textbf{22.1} & \textbf{23.4} & -- & \textbf{19.6} & \textbf{21.2} & \textbf{42.0} \\
\bottomrule
\end{tabular}}
\end{table*}

\section{Experiments}
\label{sec:experiments}

We seek to answer three questions through our experiments:
\begin{itemize}[leftmargin=*,itemsep=1pt,topsep=2pt]
  \item \textbf{Performance comparison.} In our CarlHOI framework, does steering a frozen transport policy with \carlwm{} improve its success, and how does \carlwm{} compare with existing state-of-the-art world-model planners 
  under matched simulation budgets and model sizes?
  \item \textbf{Generalization ability.} \carlwm{} is trained only with four humanoids, one frozen policy (\teamhoift{}), the table object, 2.1 m gates and 1.25--1.75\,m walls. Does it transfer, without retraining, to other unseen team sizes, obstacle geometries and placements, objects, transport policies, and to obstacles chained into a course?
  \item \textbf{Analysis.} Which components of \carlwm{} matter, and how does the team adapt its behavior to overcome obstacles?
\end{itemize}

\paragraph{Benchmark.}
We create and evaluate three cooperative transport tasks with challenging obstacles in Isaac Gym, 
as illustrated in Fig.~\ref{fig:overview}. In \emph{Obstacle
Wall}, a single wall with a span of 1.0--3.0\,m stands across the route, and the team must carry the object around it.
In \emph{Narrow Passage}, the route leads through a gate whose gap of 2.1 or 2.0\,m leaves 10\,cm or no clearance for a
2\,m round table. In the 13\,m \emph{Multi-Obstacle Course}, two offset walls are followed by a 2.0\,m gate, a
combination never seen in training. Each task is run with three objects: round table (2\,m  diameter and 78\,kg) used for training, and
unseen bed and sofa ($1.8\times1.4$\,m and $1.8\times1.2$\,m, both 50\,kg). An episode succeeds when the
object's center comes within 3\,cm of the target. We make fair comparisons with matched episodes.

\paragraph{Transport policies.}
Planners act through the policy's goal input and never change its weights. We use frozen policies:
(1) \teamhoi{}~\cite{lionar2026teamhoi}; (2) \teamhoift{}, 
fine-tuned from \teamhoi{} for target arrival with four humanoids in the 2.1\,m \emph{Narrow Passage} and subsequently used to collect all \carlwm{} training data; and (3) CooHOI~\cite{gao2024coohoi}, a separately
trained cooperative-transport policy. Planning on CooHOI is therefore zero-shot transfer to a policy
that \carlwm{} has never seen.

\paragraph{World model baselines.}
We compare against (1) each policy alone, without a planner; (2) \tdmpc{}~\cite{hansen2024tdmpc2}, a single-agent latent world
model; and (3) \seqwm{}~\cite{zhao2026seqwm}, a multi-agent world model with
sequential per-agent planning. For a fair comparison, we adapt them to the interface of \carlwm{}: they steer the same frozen policy through temporary aim points every 0.5\,s, choosing continuous rotations and rescalings within the range of our command library. \tdmpc{} issues one aim point shared by the team, and \seqwm{}, which treats each humanoid as an agent, issues one per humanoid.

\paragraph{Metrics.}
We report two metrics. \emph{Success Rate} is the fraction of episodes 
in which the object successfully reaches the target before the time limit, without a humanoid falling or the object being dropped. \emph{Time to arrival} is the simulated time from the start of the episode to arrival, averaged over all successful episodes of a
task. See results for other metrics in Supplementary Material,
Sec.~\ref{sec:supp-metrics}.

\begin{figure}
    \centering
    \includegraphics[width=0.87\columnwidth]{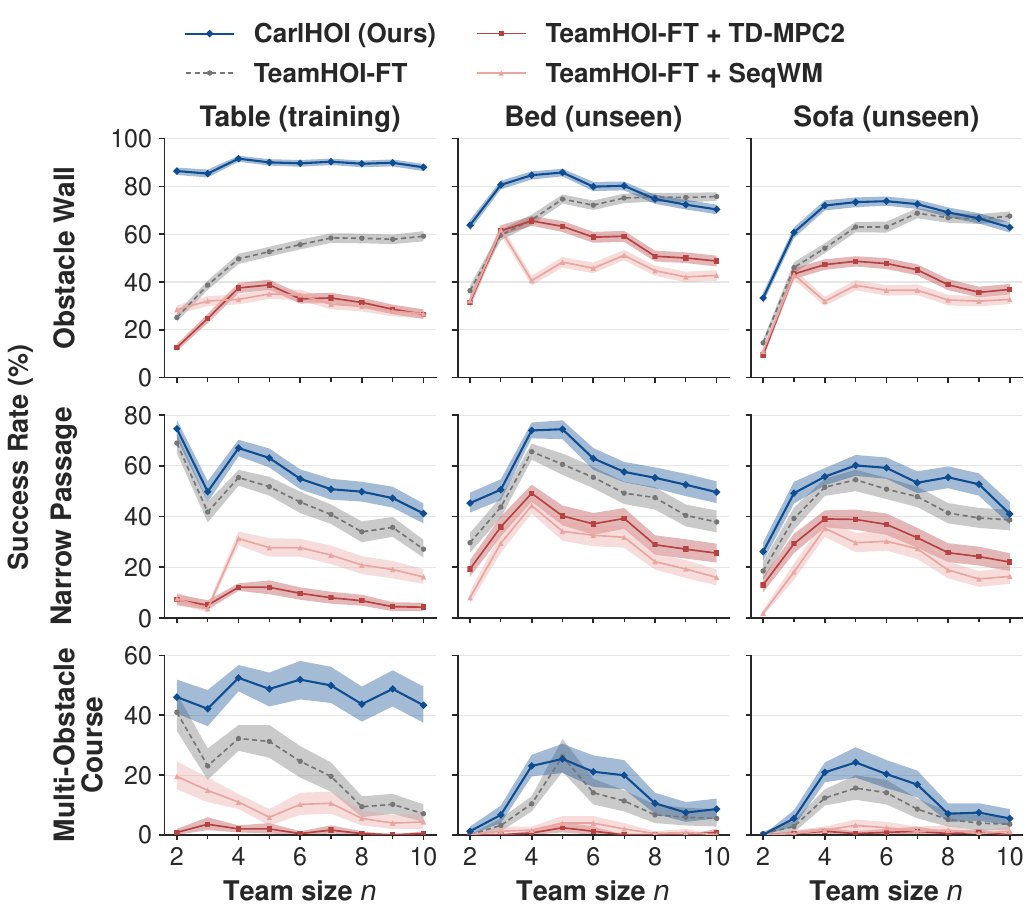}
    \caption{\textbf{Generalization to unseen objects across team sizes and tasks.} Success rate of TeamHOI-FT alone and with each world-model planner. Band: 95\% bootstrap confidence interval.}
    \label{fig:objects_teamsize_task}
\end{figure}

\section{Results}
\label{sec:Results}

\subsection{Comparison to existing methods}
\paragraph{Success rate.}
\Cref{tab:main-table-object} reports the success rate when transporting the table for every policy, planner, task, and team size.
\carlwm{} improves all three policies on all three tasks. Averaged over team sizes, it raises the success rate of
\teamhoift{} from 50.6\% to 89.0\% on the Obstacle Wall, from 44.6\% to 55.4\% through the Narrow Passage, and from
22.0\% to 47.5\% on the Multi-Obstacle Course. It achieves the highest success rate in every mean column and in all
but two combinations of policy, task and team size.

The world-model baselines, trained with 
matched simulation budgets and model sizes, \emph{lower} the success rate
of the policy they steer in all mean entries. With \teamhoift{}, \tdmpc{} and \seqwm{} reach 29.6\% and 30.9\% on
the Obstacle Wall, 7.8\% and 19.9\% through the Narrow Passage, and 1.2\% and 9.5\% on the course, all below the
policy alone. Steering a frozen policy through aim points is therefore not beneficial by itself; the gain comes from
how \carlwm{} chooses them.

\begin{figure}[t]
    \centering    \includegraphics[width=0.9\columnwidth]{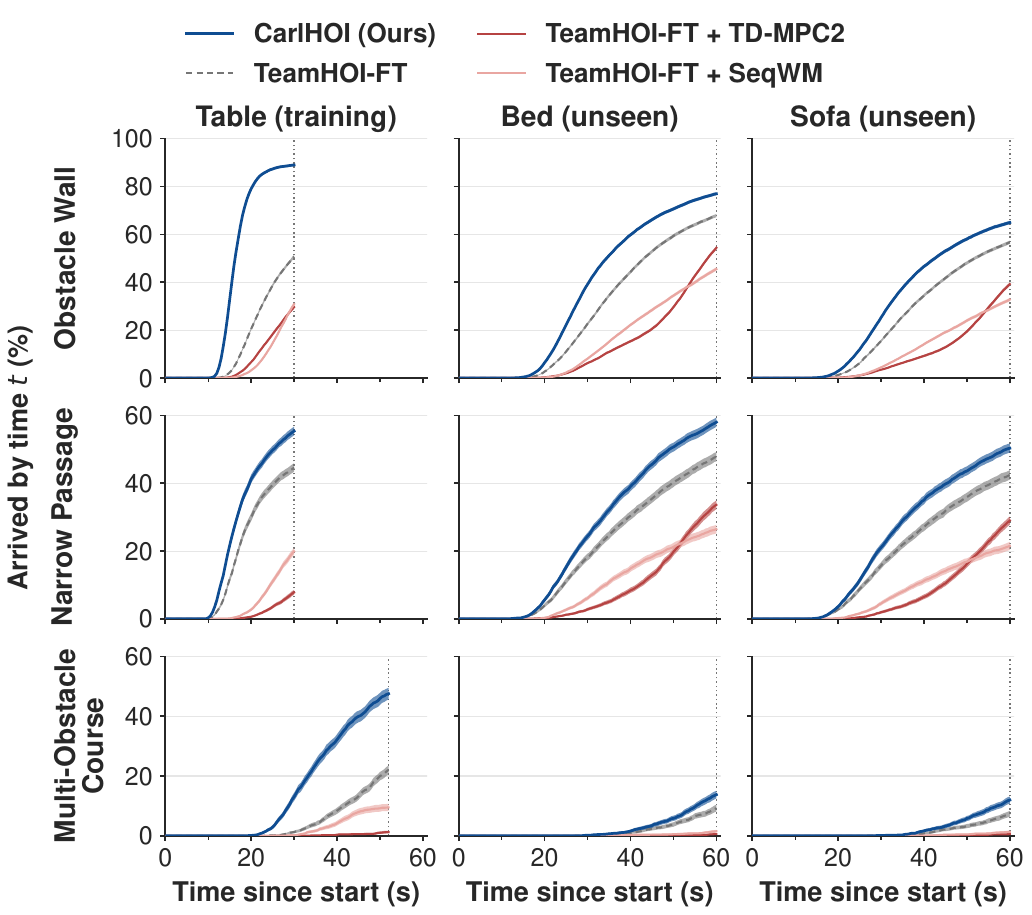}
    \caption{\textbf{Cumulative arrival rates across tasks and objects.}
Curves show the percentage of episodes reaching the target by time $t$, averaged over team sizes $n=2,\ldots,10$.
Vertical dotted lines mark deadlines: 30\,s for table wall/passage tasks, 52\,s for the table course, and 60\,s for bed/sofa tasks.}
    \label{fig:arrival_time}
\end{figure}

\paragraph{Time to arrival.}
Among successful episodes, 
CarlHOI achieves the shortest arrival time in all nine policy--task means of
\cref{tab:main-table-object}. On the Obstacle Wall, it shortens the time to arrival from 21.6 to 16.4\,s with
\teamhoift{} and from 21.1 to 15.5\,s with \teamhoi{}, whereas both baselines are slower than the policy alone. Since
this time is conditional on success, \cref{fig:arrival_time} shows the share of all episodes that have arrived by
each time. The curve of \carlwm{} never falls below those of the policy alone and of both baselines,
so it completes more episodes and completes them earlier.

\begin{figure}[t]
    \centering
    \includegraphics[width=0.9\columnwidth]{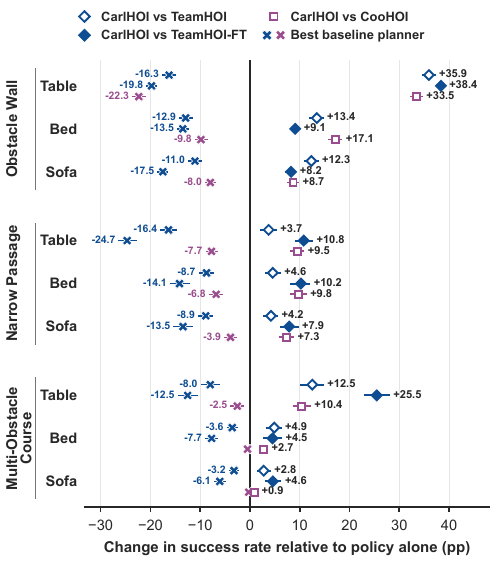}
    \caption{\textbf{Change in success rate over the policy alone} (pp: percentage points, averaged over $n=2,\dots,10$). Diamonds/squares: \ours{}; $\times$: best of \tdmpc{} and \seqwm{} on the same policy. Planners are trained only on
  the table with $n=4$; all other objects, team sizes and policies are zero-shot.}
    \label{fig:gain}
\end{figure}

\subsection{Generalization Ability}

\carlwm{} is trained exclusively with four humanoids, \teamhoift{}, the table, 2.1\,m passages and 1.25--1.75\,m walls. \textbf{All other
settings are zero-shot: other team sizes, 1.0\,m, 2.0\,m, 2.5\,m and 3.0\,m walls, the 2.0\,m gate, the 13\,m course, two other policies, and two other objects.} All the detailed results on generalization performance are reported in Supplementary Material,
Sec.~\ref{sec:supp-results}.

\paragraph{Team sizes.}
\carlwm{} improves \teamhoift{} at every team size from 2 to 10 on all three tasks (\cref{tab:main-table-object}).
The gain is largest where the policy degrades most: on the course, the policy alone drops from 41.0\% at $n=2$ to
7.0\% at $n=10$, while \carlwm{} stays between 42\% and 53\%, so its gain grows from 5.1 to 36.4 percentage points.

\paragraph{Tasks.} \carlwm{} transfers to obstacles outside its training range. As the wall widens from 1.0 to 3.0\,m, the success rate
of \teamhoift{} alone falls from 79.6\% to 16.0\%, while with \carlwm{} it stays between 
84.5\% and 93.1\%; on the
widest, unseen 3.0\,m wall, it rises from 16.0\% to 84.5\% (averaged over team sizes). Through the unseen 2.0\,m gate,
which leaves no clearance for the table, it improves from 29.6\% to 39.2\%. On the course task, it doubles the success rate on both layouts (23.9\% to 46.7\% and 20.2\% to
48.3\%).

\paragraph{Policies.}
\carlwm{}, without retraining, steers policies that it has never seen. It raises the mean success rate of
\teamhoi{} from 44.4\% to 80.3\% and that of CooHOI from 42.8\% to 76.3\% on the Obstacle Wall, and improves both on
the Narrow Passage ($+3.7$ and $+9.5$ points) and on the course ($+12.5$ and $+10.4$ points).

\paragraph{Objects.}
\Cref{fig:gain} summarizes the change in success rate over the policy alone for the bed and the sofa, which
differ from the round table in shape and mass. \carlwm{} improves the policy in all combinations of policy, object
and task, by 0.9 to 17.1 percentage points. The best baseline planner improves none
of them: it lowers the success rate substantially. The gains on the unseen objects are smaller than on the
table, for example, $+9.1$ and $+8.2$ points on the Obstacle Wall with \teamhoift{}, compared with $+38.4$ with the table.
\Cref{fig:objects_teamsize_task} breaks these gains down by team size. On the passage task, \carlwm{} improves the
policy at every team size for both objects. On the wall task, the gain shrinks as the team grows and turns
negative for the largest teams ($-0.8$ to $-5.4$ points for the bed at $n\geq8$, $-4.8$ for the sofa at $n=10$). The course task remains difficult for the bed and the sofa under every method.


\begin{table}[t]
\centering
\caption{\textbf{Component ablation.} Results for the table object, four humanoids and \teamhoift{}. Narrow
Passage (P): 2.1 and 2.0\,m gates; Multi-Obstacle Course (M): one 13\,m layout. $\Delta$:
change from the full model in pp (percentage points), paired by episode.
}
\label{tab:ablation}
\small
\setlength{\tabcolsep}{4pt}
\begin{tabular*}{\columnwidth}{@{\extracolsep{\fill}}lcccc@{}}
\toprule
 & \multicolumn{4}{c}{Success Rate (\%)} \\
\cmidrule(l){2-5}
Variant & P & $\Delta$ & M & $\Delta$ \\
\midrule
\textbf{Full CarlHOI} & \textbf{68.0} & --- & \textbf{55.1} & --- \\
\midrule
No counterfactual anchoring & 64.6 & $-3.4$ & 34.8 & $-20.3$ \\
No long-horizon value & 59.2 & $-8.8$ & 34.8 & $-20.3$ \\
Flat encoder (MLP) & 63.7 & $-4.3$ & 50.0 & $-5.1$ \\
No ranking loss & 60.7 & $-7.3$ & 52.0 & $-3.1$ \\
\midrule
\teamhoift{} & 55.4 & $-12.6$ & 27.7 & $-27.4$ \\
\bottomrule
\end{tabular*}
\end{table}

\subsection{Ablation Study}

As shown in \cref{tab:ablation}, we ablate four components of CarlWM:
the relational encoder, long-horizon value, counterfactual
anchoring, and ranking loss. Every component contributes.
On the passage task, removing the long-horizon value causes
the largest drop (8.8 pp), followed by removing the ranking
loss (7.3 pp), the relational encoder (4.3 pp), and
counterfactual anchoring (3.4 pp). On the course task,
removing counterfactual anchoring or the long-horizon value
lowers success from 55.1\% to 34.8\%. Removing the relational
encoder and ranking loss reduces success by 5.1 and 3.1 pp,
respectively.

\subsection{Insights}
\label{sec:insights}

\begin{figure}
    \centering
    \includegraphics[width=0.62\columnwidth]{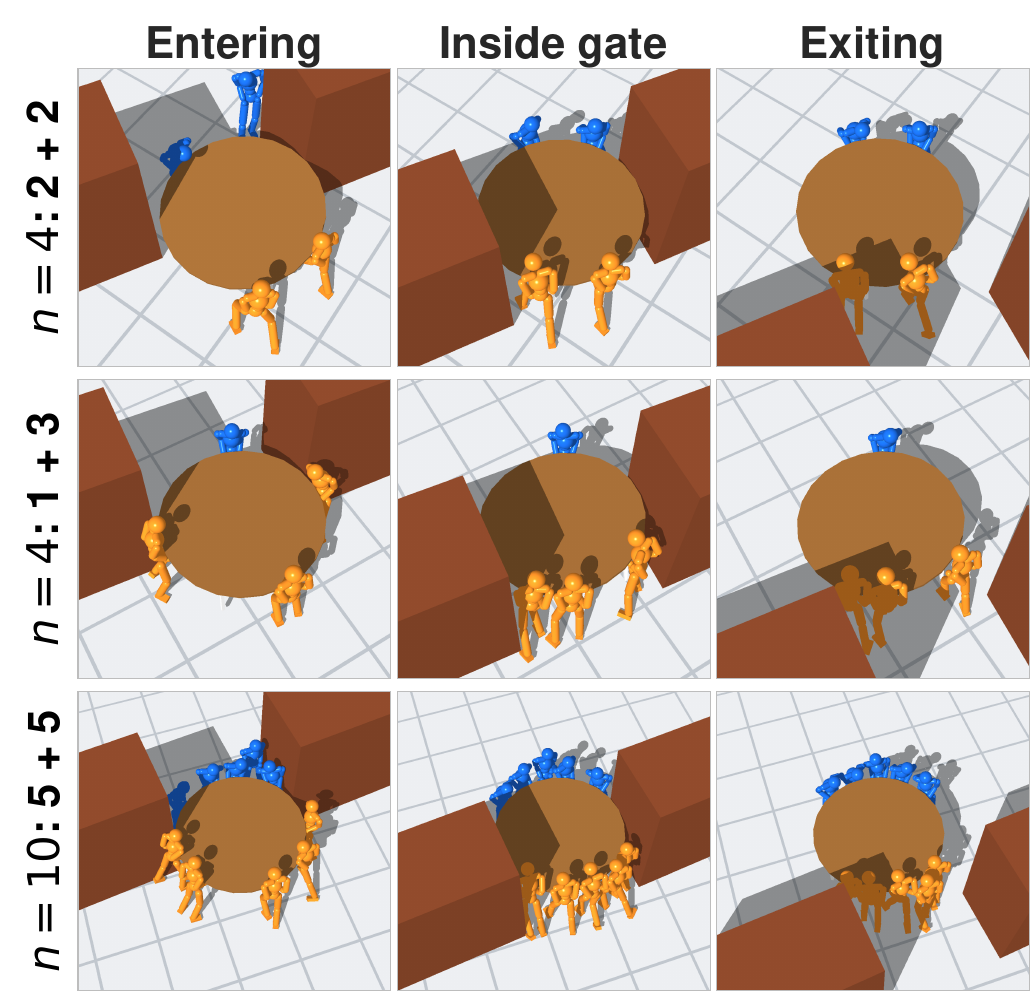}
    \caption{\textbf{Team reformation} during successful CarlHOI crossings of 2.0\,m gates (matching the table's diameter). Teams reorganize into front (blue) and rear (orange) groups: 2+2, 1+3, and 5+5. Formations of other team sizes are shown in Supplementary Material, Sec.~\ref{sec:supp-reformation}.}
    \label{fig:reformation}
\end{figure}

\paragraph{Team reformation and ``kickstart''.}
During successful passage crossings, teams often reorganize into front and rear groups while maintaining support of the table (\cref{fig:reformation}). CarlHOI guides the approach through shared waypoints; the frozen policy produces the reformation and can also exhibit it without a planner. In a few stalled episodes with one leader and three followers, the middle follower raises or alternates its legs toward the table, as if pushing it forward. We call this behavior \emph{kickstart} and report it as a qualitative observation.


\paragraph{Guidance quality and its limits.}
With ten humanoids, \carlwm{} raises \teamhoift{}'s table-course success from 7.0\% to 43.4\% without policy updates (\cref{tab:main-table-object}), whereas \tdmpc{} and \seqwm{} reduce average success. Effective waypoint selection can therefore recover existing cooperative capabilities, but shared guidance alone does not ensure improvement. Gains diminish or reverse for larger teams carrying unseen objects around walls, and multi-obstacle transport remains difficult (\cref{fig:objects_teamsize_task}). Arrival curves still rising at the deadline also suggest time-budget constraints (\cref{fig:arrival_time}). These limits motivate jointly adapting the planner and transport policy to improve guidance and execution.

\section{Conclusion}
\label{sec:conclusion}
We presented CarlHOI, a counterfactual predictive imagination framework for guiding frozen multi-humanoid transport policies through environmental constraints. Its relational world model, CarlWM, learns from counterfactual rollouts 
to predict collective responses to alternative waypoint commands, enabling long-horizon guidance for coordinated detours and team reformation. Across our benchmark of obstacle walls, narrow passages, and multi-obstacle courses, CarlHOI improves average success over the evaluated policies and world-model planners. Trained with four humanoids, a table, and one policy, it transfers without retraining to teams of two to ten, unseen objects, other policies, and unseen obstacle configurations. 
These results demonstrate the potential of predictive shared guidance to extend pretrained cooperative transport skills, supporting generalizable multi-humanoid object interaction. 

\section*{LLM Usage Statement}
OpenAI ChatGPT was used to assist with language refinement, grammatical corrections, and improvements to the clarity and organization of the manuscript.  All research contributions, methodological decisions, experimental results, and conclusions were developed and validated by the authors, who take full responsibility for the content of this paper.
{
    \small
    \bibliographystyle{ieeenat_fullname}
    \bibliography{main}

@String(CVPR= {IEEE Conf. Comput. Vis. Pattern Recog.})

@String(TOG= {ACM Trans. Graph.})

@String(CVPR  = {CVPR})

@String(TOG   = {ACM TOG})

@inproceedings{gao2024coohoi,
  title = {{CooHOI}: Learning Cooperative Human-Object Interaction with Manipulated Object Dynamics},
  author = {Gao, Jiawei and Wang, Ziqin and Xiao, Zeqi and Wang, Jingbo and Wang, Tai and Cao, Jinkun and Hu, Xiaolin and Liu, Si and Dai, Jifeng and Pang, Jiangmiao},
  booktitle = {Advances in Neural Information Processing Systems},
  volume = {37},
  year = {2024},
  url = {https://papers.nips.cc/paper_files/paper/2024/hash/918b9487f8ea4661e8ba5a02b2126658-Abstract-Conference.html}
}

@inproceedings{lionar2026teamhoi,
  title = {{TeamHOI}: Learning a Unified Policy for Cooperative Human-Object Interactions with Any Team Size},
  author = {Lionar, Stefan and Lee, Gim Hee},
  booktitle = {Proceedings of the IEEE/CVF Conference on Computer Vision and Pattern Recognition},
  pages = {37121--37132},
  year = {2026},
  url = {https://arxiv.org/abs/2603.07988}
}

@article{pandit2026decmht,
  title = {Learning Multi-Humanoid Pickup and Transport via Decentralized Object-Centric Control},
  author = {Pandit, Bikram and Gadde, Mohitvishnu S. and Shrestha, Aayam Kumar and Fern, Alan},
  journal = {arXiv preprint arXiv:2609.17824},
  year = {2026},
  url = {https://arxiv.org/abs/2609.17824}
}

@article{deng2026ensemble,
  title = {From Solo to Ensemble: A Hierarchical Framework for Composable Multi-Agent Human-Object Interaction},
  author = {Deng, Zekai and Chen, Kangyi and Shi, Ye and Wang, Jingya},
  journal = {arXiv preprint arXiv:2610.11722},
  year = {2026},
  url = {https://arxiv.org/abs/2610.11722}
}

@inproceedings{hansen2024tdmpc2,
  title = {{TD-MPC2}: Scalable, Robust World Models for Continuous Control},
  author = {Hansen, Nicklas and Su, Hao and Wang, Xiaolong},
  booktitle = {International Conference on Learning Representations},
  year = {2024},
  url = {https://arxiv.org/abs/2310.16828}
}

@article{shi2022skill,
  title={Skill-based model-based reinforcement learning},
  author={Shi, Lucy Xiaoyang and Lim, Joseph J and Lee, Youngwoon},
  journal={arXiv preprint arXiv:2207.07560},
  year={2022}
}

@inproceedings{freed2023abstract,
  title = {Learning Temporally Abstract World Models without Online Experimentation},
  author = {Freed, Benjamin and Venkatraman, Siddarth and Sartoretti, Guillaume Adrien and Schneider, Jeff and Choset, Howie},
  booktitle = {Proceedings of the 40th International Conference on Machine Learning},
  series = {Proceedings of Machine Learning Research},
  volume = {202},
  pages = {10338--10356},
  year = {2023},
  publisher = {PMLR},
  url = {https://proceedings.mlr.press/v202/freed23a.html}
}

@inproceedings{zhang2025combo,
  title = {{COMBO}: Compositional World Models for Embodied Multi-Agent Cooperation},
  author = {Zhang, Hongxin and Wang, Zeyuan and Lyu, Qiushi and Zhang, Zheyuan and Chen, Sunli and Shu, Tianmin and Dariush, Behzad and Lee, Kwonjoon and Du, Yilun and Gan, Chuang},
  booktitle = {International Conference on Learning Representations},
  year = {2025},
  url = {https://arxiv.org/abs/2404.10775}
}

@inproceedings{zhao2026seqwm,
  title = {Empowering Multi-Robot Cooperation via Sequential World Models},
  author = {Zhao, Zijie and Guo, Honglei and Chen, Shengqian and Xu, Kaixuan and Jiang, Bo and Zhu, Yuanheng and Zhao, Dongbin},
  booktitle = {International Conference on Learning Representations},
  year = {2026},
  url = {https://openreview.net/forum?id=IvUM6UwYCJ}
}

@inproceedings{battaglia2016interaction,
  title = {Interaction Networks for Learning about Objects, Relations and Physics},
  author = {Battaglia, Peter W. and Pascanu, Razvan and Lai, Matthew and Rezende, Danilo Jimenez and Kavukcuoglu, Koray},
  booktitle = {Advances in Neural Information Processing Systems},
  volume = {29},
  year = {2016},
  url = {https://papers.nips.cc/paper_files/paper/2016/hash/3147da8ab4a0437c15ef51a5cc7f2dc4-Abstract.html}
}

@article{kipf2019contrastive,
  title={Contrastive learning of structured world models},
  author={Kipf, Thomas and Van der Pol, Elise and Welling, Max},
  journal={arXiv preprint arXiv:1911.12247},
  year={2019}
}

@article{lee2026mochi,
  title={{MOCHI}: Motion Enhancement of Collaborative Human-object Interactions},
  author={Lee, Jiye and Choi, Yonghun and Won, Jungdam},
  journal={ACM Transactions on Graphics (TOG)},
  volume={45},
  number={4},
  pages={1--18},
  year={2026},
  publisher={ACM, New York, NY, USA}
}

@article{lu2025choice,
  title={CHOICE: Coordinated human-object interaction in cluttered environments for pick-and-place actions},
  author={Lu, Jintao and Zhang, He and Ye, Yuting and Shiratori, Takaaki and Starke, Sebastian and Komura, Taku},
  journal={ACM Transactions on Graphics},
  volume={45},
  number={2},
  pages={1--18},
  year={2025},
  publisher={ACM, New York, NY}
}

@article{wang2023physhoi,
  title={{PhysHOI}: Physics-based imitation of dynamic human-object interaction},
  author={Wang, Yinhuai and Lin, Jing and Zeng, Ailing and Luo, Zhengyi and Zhang, Jian and Zhang, Lei},
  journal={arXiv preprint arXiv:2312.04393},
  year={2023}
}

@inproceedings{jia2025primhoi,
  title={{Primhoi}: Compositional human-object interaction via reusable primitives},
  author={Jia, Kai and Liu, Tengyu and Pei, Mingtao and Zhu, Yixin and Huang, Siyuan},
  booktitle={Proceedings of the IEEE/CVF International Conference on Computer Vision},
  pages={11491--11501},
  year={2025}
}

@article{lurz2022spline,
  title={Spline-based path planning and reconfiguration for rigid multi-robot formations},
  author={Lurz, Henrik and Recker, Tobias and Raatz, Annika},
  journal={Procedia CIRP},
  volume={106},
  pages={174--179},
  year={2022},
  publisher={Elsevier}
}

@article{patra2025kinodynamic,
  title={Kinodynamic motion planning for collaborative object transportation by multiple mobile manipulators},
  author={Patra, Keshab and Sinha, Arpita and Guha, Anirban},
  journal={Journal of Mechanisms and Robotics},
  volume={17},
  number={12},
  pages={121003},
  year={2025},
  publisher={American Society of Mechanical Engineers}
}

@article{alvear2026cooperative,
  title={Cooperative Grasping for Collective Object Transport in Constrained Environments},
  author={Alvear, David and Turkiyyah, George and Park, Shinkyu},
  journal={IEEE Robotics and Automation Letters},
  year={2026},
  publisher={IEEE}
}

@article{peng2018deepmimic,
  title = {{DeepMimic}: Example-Guided Deep Reinforcement Learning of Physics-Based Character Skills},
  author = {Peng, Xue Bin and Abbeel, Pieter and Levine, Sergey and van de Panne, Michiel},
  journal = {ACM Transactions on Graphics},
  volume = {37},
  number = {4},
  articleno = {143},
  numpages = {14},
  year = {2018},
  doi = {10.1145/3197517.3201311},
  url = {https://www.cs.ubc.ca/~van/papers/2018-TOG-deepMimic/index.html}
}

@article{peng2021amp,
  title = {{AMP}: Adversarial Motion Priors for Stylized Physics-Based Character Control},
  author = {Peng, Xue Bin and Ma, Ze and Abbeel, Pieter and Levine, Sergey and Kanazawa, Angjoo},
  journal = {ACM Transactions on Graphics},
  volume = {40},
  number = {4},
  articleno = {144},
  numpages = {20},
  year = {2021},
  doi = {10.1145/3450626.3459670},
  url = {https://xbpeng.github.io/projects/AMP/}
}

@inproceedings{xu2025intermimic,
  title = {{InterMimic}: Towards Universal Whole-Body Control for Physics-Based Human-Object Interactions},
  author = {Xu, Sirui and Ling, Hung Yu and Wang, Yu-Xiong and Gui, Liang-Yan},
  booktitle = {Proceedings of the IEEE/CVF Conference on Computer Vision and Pattern Recognition},
  pages = {12266--12277},
  year = {2025},
  url = {https://openaccess.thecvf.com/content/CVPR2025/html/Xu_InterMimic_Towards_Universal_Whole-Body_Control_for_Physics-Based_Human-Object_Interactions_CVPR_2025_paper.html}
}

@article{merel2020catch,
  title = {Catch \& Carry: Reusable Neural Controllers for Vision-Guided Whole-Body Tasks},
  author = {Merel, Josh and Tunyasuvunakool, Saran and Ahuja, Arun and Tassa, Yuval and Hasenclever, Leonard and Pham, Vu and Erez, Tom and Wayne, Greg and Heess, Nicolas},
  journal = {ACM Transactions on Graphics},
  volume = {39},
  number = {4},
  articleno = {39},
  numpages = {14},
  year = {2020},
  doi = {10.1145/3386569.3392474},
  url = {https://arxiv.org/abs/1911.06636}
}

@inproceedings{sanchezgonzalez2020gns,
  title = {Learning to Simulate Complex Physics with Graph Networks},
  author = {Sanchez-Gonzalez, Alvaro and Godwin, Jonathan and Pfaff, Tobias and Ying, Rex and Leskovec, Jure and Battaglia, Peter},
  booktitle = {Proceedings of the 37th International Conference on Machine Learning},
  series = {Proceedings of Machine Learning Research},
  volume = {119},
  pages = {8459--8468},
  publisher = {PMLR},
  year = {2020},
  url = {https://proceedings.mlr.press/v119/sanchez-gonzalez20a.html}
}

@article{peng2022ase,
  title     = {{ASE}: Large-Scale Reusable Adversarial Skill Embeddings
               for Physically Simulated Characters},
  author    = {Peng, Xue Bin and Guo, Yunrong and Halper, Lina
               and Levine, Sergey and Fidler, Sanja},
  journal   = {ACM Transactions on Graphics},
  volume    = {41},
  number    = {4},
  articleno = {94},
  year      = {2022},
  url       = {https://xbpeng.github.io/projects/ASE/index.html}
}

@inproceedings{hansen2022tdmpc,
  title     = {Temporal Difference Learning for Model Predictive Control},
  author    = {Hansen, Nicklas A. and Su, Hao and Wang, Xiaolong},
  booktitle = {Proceedings of the 39th International Conference
               on Machine Learning},
  series    = {Proceedings of Machine Learning Research},
  volume    = {162},
  pages     = {8387--8406},
  publisher = {PMLR},
  year      = {2022},
  url       = {https://proceedings.mlr.press/v162/hansen22a.html}
}

@inproceedings{wang2025skillmimic,
  title={{SkillMimic}: Learning basketball interaction skills from demonstrations},
  author={Wang, Yinhuai and Zhao, Qihan and Yu, Runyi and Tsui, Hok Wai and Zeng, Ailing and Lin, Jing and Luo, Zhengyi and Yu, Jiwen and Li, Xiu and Chen, Qifeng and others},
  booktitle={2025 IEEE/CVF Conference on Computer Vision and Pattern Recognition (CVPR)},
  pages={17540--17549},
  year={2025},
  organization={IEEE}
}

@article{sharma2019dynamics,
  title={Dynamics-aware unsupervised discovery of skills},
  author={Sharma, Archit and Gu, Shixiang and Levine, Sergey and Kumar, Vikash and Hausman, Karol},
  journal={arXiv preprint arXiv:1907.01657},
  year={2019}
}

@article{tessler2024maskedmimic,
  title={{MaskedMimic}: Unified physics-based character control through masked motion inpainting},
  author={Tessler, Chen and Guo, Yunrong and Nabati, Ofir and Chechik, Gal and Peng, Xue Bin},
  journal={ACM Transactions On Graphics (TOG)},
  volume={43},
  number={6},
  pages={1--21},
  year={2024},
  publisher={ACM, New York, NY, USA}
}

@article{lin2025reflex,
  title={{REFLEX}: Metacognitive Reasoning for Reflective Zero-Shot Robotic Planning with Large Language Models},
  author={Lin, Wenjie and Wei-Kocsis, Jin and Zhang, Jiansong and Min, Byung-Cheol and Gan, Dongming and Asunda, Paul and Athinarayanan, Ragu},
  journal={arXiv preprint arXiv:2505.14899},
  year={2025}
}

@inproceedings{kipf2018neural,
  title={Neural relational inference for interacting systems},
  author={Kipf, Thomas and Fetaya, Ethan and Wang, Kuan-Chieh and Welling, Max and Zemel, Richard},
  booktitle={International conference on machine learning},
  pages={2688--2697},
  year={2018},
  organization={Pmlr}
}

@article{li2018learning,
  title={Learning particle dynamics for manipulating rigid bodies, deformable objects, and fluids},
  author={Li, Yunzhu and Wu, Jiajun and Tedrake, Russ and Tenenbaum, Joshua B and Torralba, Antonio},
  journal={arXiv preprint arXiv:1810.01566},
  year={2018}
}

@inproceedings{NEURIPS2024_215aeb07,
 author = {Xu, Xinyu and Zhang, Yizheng and Li, Yong-Lu and Han, Lei and Lu, Cewu},
 booktitle = {Advances in Neural Information Processing Systems},
 doi = {10.52202/079017-0591},
 editor = {A. Globerson and L. Mackey and D. Belgrave and A. Fan and U. Paquet and J. Tomczak and C. Zhang},
 pages = {18633--18659},
 publisher = {Curran Associates, Inc.},
 title = {HumanVLA: Towards Vision-Language Directed Object Rearrangement by Physical Humanoid},
 url = {https://proceedings.neurips.cc/paper_files/paper/2024/file/215aeb07b5996c969c0123c3c6ee8f54-Paper-Conference.pdf},
 volume = {37},
 year = {2024}
}

@article{peng2019mcp,
  title={{MCP}: Learning composable hierarchical control with multiplicative compositional policies},
  author={Peng, Xue Bin and Chang, Michael and Zhang, Grace and Abbeel, Pieter and Levine, Sergey},
  journal={Advances in neural information processing systems},
  volume={32},
  year={2019}
}
}


\clearpage
\setcounter{page}{1}
\maketitlesupplementary

\section{Additional results}
\label{sec:supp-results}

\subsection{Results per object, task, policy and team size}
\label{sec:supp-detailed}
\Cref{tab:supp-bed,tab:supp-sofa} repeat \cref{tab:main-table-object} for the unseen bed and sofa, with every policy,
planner, task and team size. \Cref{tab:supp-subtask-table,tab:supp-subtask-bed,tab:supp-subtask-sofa} report the
success rate for each individual obstacle task (Obstacle Wall, Narrow Passage and Multi-Obstacle Course), averaged over team sizes. \Cref{tab:supp-subtask-table,tab:supp-subtask-bed,tab:supp-subtask-sofa} show only testing wall spans. The aggregate wall results for TeamHOI-FT in \cref{tab:main-table-object,tab:supp-bed,tab:supp-sofa} additionally include the 1.25, 1.5, and 1.75\,m spans. \carlwm{} gives the highest success rate (or ties for it) for every policy, object and individual obstacle, including the
testing layouts of 1.0, 2.0, 2.5 and 3.0\,m walls, the 2.0\,m gate and both 13\,m course tasks. Its advantage is largest on the
widest walls, where the frozen policies fail most often.
\begin{table*}[t]
\centering
\caption{\textbf{Additional metrics and final distance} for the table object, averaged over $n=2,\dots,10$ as in \cref{tab:main-table-object}. Supp.: the team stopped supporting the object; N.A.: not arrived (time limit or route termination without arrival); all as \% of episodes. Dist.: mean distance from the object center to the target at the end of the episode (m), over all episodes. Bold: lowest value per policy and column.}
\label{tab:supp-metrics}
\footnotesize
\setlength{\tabcolsep}{2.6pt}
\begin{tabular}{@{}ll@{\hspace{6pt}}ccc@{\hspace{6pt}}ccc@{\hspace{6pt}}ccc@{}}
\toprule
 & & \multicolumn{3}{c}{Obstacle Wall} & \multicolumn{3}{c}{Narrow Passage} & \multicolumn{3}{c}{Multi-Obstacle Course} \\
\cmidrule(lr){3-5}\cmidrule(lr){6-8}\cmidrule(l){9-11}
Policy & Planner & Supp.\,$\downarrow$ & N.A.\,$\downarrow$ & Dist.\,$\downarrow$ & Supp.\,$\downarrow$ & N.A.\,$\downarrow$ & Dist.\,$\downarrow$ & Supp.\,$\downarrow$ & N.A.\,$\downarrow$ & Dist.\,$\downarrow$ \\
\midrule
\multirow{4}{*}{\teamhoi{}} & Policy alone & 9.2 & 36.3 & 2.58 & 23.8 & 15.2 & 3.77 & 0.1 & 59.5 & 4.21 \\
 & \tdmpc{} & 2.9 & 54.9 & 2.00 & \textbf{22.9} & 39.3 & 4.55 & 0.2 & 75.0 & 4.31 \\
 & \seqwm{} & 7.8 & 52.2 & 2.53 & 24.7 & 30.1 & 4.31 & \textbf{0.0} & 69.1 & \textbf{2.95} \\
 & \carlwm{} (Ours) & \textbf{0.2} & \textbf{3.3} & \textbf{0.87} & 27.0 & \textbf{8.2} & \textbf{3.54} & \textbf{0.0} & \textbf{38.7} & 3.28 \\
\cmidrule(l){2-11}
\multirow{4}{*}{\teamhoift{}} & Policy alone & 3.5 & 40.4 & 2.25 & 31.4 & 14.9 & 2.48 & 0.3 & 61.0 & 4.41 \\
 & \tdmpc{} & 1.1 & 61.0 & 1.54 & 33.3 & 48.1 & 3.73 & 0.2 & 83.0 & 3.99 \\
 & \seqwm{} & 3.1 & 59.1 & 2.06 & 35.6 & 34.9 & 3.18 & \textbf{0.0} & 75.8 & 2.58 \\
 & \carlwm{} (Ours) & \textbf{0.0} & \textbf{2.8} & \textbf{0.55} & \textbf{30.4} & \textbf{5.8} & \textbf{1.95} & \textbf{0.0} & \textbf{32.1} & \textbf{2.37} \\
\bottomrule
\end{tabular}
\end{table*}

\begin{table*}[t]
\centering
\caption{\textbf{Carrying the unseen bed across tasks and team sizes.} Layout and aggregation as in \cref{tab:main-table-object}: W: Obstacle Wall (\teamhoi{} and \coohoi{}: 1.0 and 2.0--3.0\,m), P: Narrow Passage (2.1 and 2.0\,m gates), M: Multi-Obstacle Course (two layouts). $^\dagger$Planners are trained with $n=4$, \teamhoift{} and the table only. Time: mean arrival time of successful episodes (limit 60\,s); --: no successful episode.}
\label{tab:supp-bed}
\setlength{\tabcolsep}{1.3pt}
\resizebox{\textwidth}{!}{%
\begin{tabular}{@{}ll@{\hspace{5pt}}ccc@{\hspace{5pt}}ccc@{\hspace{5pt}}ccc@{\hspace{5pt}}ccc@{\hspace{5pt}}ccc@{\hspace{5pt}}ccc@{\hspace{5pt}}ccc@{\hspace{5pt}}ccc@{\hspace{5pt}}ccc@{\hspace{5pt}}ccc@{}}
\toprule
 & & \multicolumn{27}{c}{Team size $n$} & \\
\cmidrule(lr){3-29}
 & & \multicolumn{3}{c}{2} & \multicolumn{3}{c}{3} & \multicolumn{3}{c}{4$^\dagger$} & \multicolumn{3}{c}{5} & \multicolumn{3}{c}{6} & \multicolumn{3}{c}{7} & \multicolumn{3}{c}{8} & \multicolumn{3}{c}{9} & \multicolumn{3}{c}{10} & \multicolumn{3}{c}{Mean} \\
\cmidrule(lr){3-5}\cmidrule(lr){6-8}\cmidrule(lr){9-11}\cmidrule(lr){12-14}\cmidrule(lr){15-17}\cmidrule(lr){18-20}\cmidrule(lr){21-23}\cmidrule(lr){24-26}\cmidrule(lr){27-29}\cmidrule(lr){30-32}
Policy & Planner & W & P & M & W & P & M & W & P & M & W & P & M & W & P & M & W & P & M & W & P & M & W & P & M & W & P & M & W & P & M \\
\midrule
\multicolumn{32}{@{}l}{\textit{Success rate (\%) $\uparrow$}} \\
\midrule
\multirow{4}{*}{\teamhoi{}} & Policy alone & 19.5 & 7.8 & 0.0 & 46.6 & 24.6 & 1.2 & 48.7 & 42.2 & 3.5 & 53.7 & 35.2 & 7.4 & 51.6 & 31.2 & 5.5 & 50.6 & 25.0 & 8.6 & 46.7 & 18.4 & 3.9 & 48.6 & 14.3 & 5.5 & 49.8 & \textbf{15.0} & 5.9 & 46.2 & 23.7 & 4.6 \\
 & \tdmpc{} & 17.2 & 4.5 & 0.0 & 43.9 & 21.7 & 0.0 & 47.9 & 27.2 & 0.0 & 47.9 & 24.4 & 0.0 & 38.9 & 18.0 & 0.8 & 32.5 & 15.2 & 0.4 & 24.5 & 8.4 & 0.8 & 23.2 & 8.8 & 0.8 & 23.3 & 6.8 & 1.2 & 33.3 & 15.0 & 0.4 \\
 & \seqwm{} & 18.5 & 0.4 & 0.0 & 46.1 & 13.7 & 0.0 & 30.4 & 29.1 & 1.6 & 32.8 & 22.3 & 1.6 & 27.3 & 18.0 & 1.2 & 25.4 & 13.5 & 1.6 & 19.6 & 7.4 & 1.6 & 20.8 & 5.9 & 0.4 & 24.5 & 6.1 & 1.6 & 27.3 & 12.9 & 1.0 \\
 & \textbf{\carlwm{} (Ours)} & \textbf{36.7} & \textbf{18.4} & \textbf{0.4} & \textbf{65.9} & \textbf{33.4} & \textbf{7.4} & \textbf{74.6} & \textbf{48.5} & \textbf{17.4} & \textbf{69.8} & \textbf{42.8} & \textbf{13.3} & \textbf{66.9} & \textbf{33.2} & \textbf{12.1} & \textbf{63.0} & \textbf{28.5} & \textbf{10.2} & \textbf{55.8} & \textbf{18.9} & \textbf{10.5} & \textbf{52.0} & \textbf{16.2} & \textbf{6.6} & \textbf{52.0} & 14.8 & \textbf{7.4} & \textbf{59.6} & \textbf{28.3} & \textbf{9.5} \\
\cmidrule(l){2-32}
\multirow{4}{*}{\teamhoift{}$^\dagger$} & Policy alone & 36.4 & 29.7 & 0.0 & 59.6 & 43.8 & 3.1 & 66.0 & 65.6 & 10.4 & 74.7 & 60.5 & \textbf{26.2} & 72.2 & 55.5 & 14.1 & 75.2 & 49.2 & 11.3 & \textbf{75.5} & 47.5 & 6.6 & \textbf{75.4} & 40.4 & 5.9 & \textbf{75.8} & 37.9 & 5.5 & 67.9 & 47.8 & 9.2 \\
 & \tdmpc{} & 31.5 & 19.3 & 0.0 & 61.4 & 35.9 & 0.0 & 65.6 & 49.2 & 0.6 & 63.3 & 40.2 & 2.3 & 58.8 & 37.1 & 1.2 & 59.2 & 39.3 & 0.0 & 50.8 & 28.9 & 0.0 & 50.1 & 27.1 & 0.0 & 48.8 & 25.6 & 0.8 & 54.4 & 33.6 & 0.5 \\
 & \seqwm{} & 32.2 & 8.0 & 0.0 & 61.9 & 29.5 & 1.2 & 40.6 & 44.3 & 1.6 & 48.3 & 34.2 & 3.9 & 45.8 & 32.6 & 3.9 & 51.2 & 31.8 & 2.0 & 44.7 & 22.3 & 0.4 & 42.1 & 19.3 & 0.8 & 42.7 & 16.0 & 0.0 & 45.5 & 26.5 & 1.5 \\
 & \textbf{\carlwm{} (Ours)} & \textbf{63.8} & \textbf{45.3} & \textbf{1.2} & \textbf{80.6} & \textbf{50.6} & \textbf{6.6} & \textbf{84.7} & \textbf{73.9} & \textbf{23.0} & \textbf{85.8} & \textbf{74.4} & 25.4 & \textbf{79.9} & \textbf{62.9} & \textbf{21.1} & \textbf{80.2} & \textbf{57.6} & \textbf{19.9} & 74.7 & \textbf{55.3} & \textbf{10.5} & 72.4 & \textbf{52.5} & \textbf{7.4} & 70.4 & \textbf{49.6} & \textbf{8.6} & \textbf{77.0} & \textbf{58.0} & \textbf{13.8} \\
\cmidrule(l){2-32}
\multirow{4}{*}{\coohoi{}} & Policy alone & 23.7 & 38.9 & 0.0 & 15.3 & 5.9 & 0.0 & 44.8 & 32.6 & 0.0 & 8.0 & 5.5 & 0.8 & 43.3 & 10.7 & 0.0 & 38.9 & 18.0 & 0.4 & 45.0 & 19.5 & 2.0 & 38.8 & 15.2 & 2.7 & 43.8 & 22.3 & 0.4 & 33.5 & 18.7 & 0.7 \\
 & \tdmpc{} & 27.1 & 29.7 & 0.0 & 9.7 & 3.5 & 0.0 & 36.4 & 20.5 & 0.0 & 4.2 & 3.7 & 0.8 & 28.6 & 4.7 & 0.0 & 21.4 & 11.5 & 0.4 & 29.6 & 10.4 & 0.0 & 24.3 & 8.0 & 0.4 & 31.7 & 14.3 & 0.0 & 23.7 & 11.8 & 0.2 \\
 & \seqwm{} & 31.0 & 15.6 & 0.0 & 11.9 & 3.3 & 0.0 & 18.8 & 26.1 & 0.0 & 5.7 & 2.9 & 0.8 & 25.5 & 7.2 & 0.0 & 23.8 & 11.1 & 0.0 & 25.5 & 13.7 & 0.8 & 28.1 & 11.5 & 0.0 & 26.3 & 15.8 & 0.4 & 21.8 & 11.9 & 0.2 \\
 & \textbf{\carlwm{} (Ours)} & \textbf{67.2} & \textbf{60.4} & 0.0 & \textbf{19.3} & \textbf{8.4} & 0.0 & \textbf{69.4} & \textbf{49.1} & \textbf{6.6} & \textbf{10.1} & \textbf{7.2} & \textbf{2.3} & \textbf{57.6} & \textbf{17.2} & \textbf{1.6} & \textbf{54.9} & \textbf{27.7} & \textbf{4.3} & \textbf{62.5} & \textbf{34.8} & \textbf{7.0} & \textbf{54.4} & \textbf{20.1} & \textbf{3.1} & \textbf{60.3} & \textbf{31.6} & \textbf{5.5} & \textbf{50.6} & \textbf{28.5} & \textbf{3.4} \\
\midrule
\multicolumn{32}{@{}l}{\textit{Time to arrival (s), successful episodes only $\downarrow$}} \\
\midrule
\multirow{4}{*}{\teamhoi{}} & Policy alone & 47.0 & \textbf{37.6} & -- & 43.7 & \textbf{40.3} & \textbf{48.0} & 40.0 & 34.5 & 51.4 & 37.3 & 36.6 & 51.4 & 36.7 & 35.9 & 46.9 & 37.6 & 34.7 & \textbf{47.6} & 37.1 & 37.8 & 47.0 & 38.3 & 34.9 & 48.9 & 36.5 & \textbf{34.4} & 47.7 & 39.4 & \textbf{36.3} & 48.6 \\
 & \tdmpc{} & 49.9 & 47.9 & -- & 48.0 & 48.7 & -- & 46.5 & 44.4 & -- & 44.9 & 47.1 & -- & 46.2 & 47.1 & 56.5 & 45.4 & 45.2 & 57.1 & 45.5 & 43.9 & 45.9 & 45.0 & 46.4 & 54.4 & 44.5 & 44.7 & 54.9 & 46.2 & 46.2 & 53.7 \\
 & \seqwm{} & 48.8 & 56.0 & -- & 44.6 & 46.4 & -- & 43.8 & 38.5 & 53.7 & 40.3 & 39.4 & 53.3 & 39.7 & 38.7 & \textbf{38.9} & 40.3 & 37.7 & 57.2 & 41.8 & 39.0 & \textbf{44.2} & 39.5 & 39.6 & \textbf{38.5} & 39.1 & 35.6 & \textbf{45.3} & 42.0 & 41.2 & \textbf{47.3} \\
 & \textbf{\carlwm{} (Ours)} & \textbf{42.2} & 43.3 & 50.1 & \textbf{37.6} & 43.3 & 51.2 & \textbf{33.6} & \textbf{34.4} & \textbf{51.1} & \textbf{33.3} & \textbf{34.8} & \textbf{50.0} & \textbf{33.6} & \textbf{34.8} & 49.1 & \textbf{34.5} & \textbf{33.8} & 47.8 & \textbf{35.2} & \textbf{36.1} & 48.4 & \textbf{34.4} & \textbf{34.4} & 47.7 & \textbf{34.4} & 36.8 & 46.5 & \textbf{35.4} & 36.9 & 49.1 \\
\cmidrule(l){2-32}
\multirow{4}{*}{\teamhoift{}$^\dagger$} & Policy alone & 47.9 & 42.9 & -- & 39.3 & \textbf{39.5} & 53.4 & 36.9 & 32.6 & 52.0 & 33.1 & 33.6 & \textbf{47.8} & 33.4 & 34.6 & 49.3 & 33.6 & 34.9 & 49.3 & 35.6 & 36.7 & 52.3 & 35.5 & 34.7 & 54.9 & 35.7 & 35.4 & \textbf{51.2} & 36.8 & 36.1 & 51.3 \\
 & \tdmpc{} & 50.7 & 48.6 & -- & 47.1 & 47.2 & -- & 45.1 & 44.6 & 52.6 & 43.6 & 44.6 & 53.6 & 45.1 & 46.0 & 57.6 & 45.0 & 46.0 & -- & 45.8 & 45.3 & -- & 46.1 & 44.7 & -- & 46.8 & 48.0 & 58.4 & 46.1 & 46.1 & 55.5 \\
 & \seqwm{} & 48.9 & 48.9 & -- & 42.9 & 46.1 & \textbf{52.1} & 39.7 & 37.4 & 53.9 & 37.8 & 37.3 & 53.6 & 38.6 & 37.3 & \textbf{47.7} & 39.0 & 38.0 & \textbf{44.8} & 40.7 & 38.2 & 57.8 & 41.5 & 36.6 & 56.0 & 41.9 & 37.4 & -- & 41.2 & 39.7 & 52.3 \\
 & \textbf{\carlwm{} (Ours)} & \textbf{43.1} & \textbf{42.3} & 45.2 & \textbf{35.4} & 39.6 & 53.2 & \textbf{29.6} & \textbf{31.6} & \textbf{50.2} & \textbf{27.9} & \textbf{31.8} & 48.0 & \textbf{28.8} & \textbf{32.4} & 48.8 & \textbf{30.4} & \textbf{32.0} & 48.5 & \textbf{32.5} & \textbf{35.5} & \textbf{52.1} & \textbf{32.4} & \textbf{33.4} & \textbf{51.6} & \textbf{32.0} & \textbf{34.4} & 52.4 & \textbf{32.4} & \textbf{34.8} & \textbf{50.0} \\
\cmidrule(l){2-32}
\multirow{4}{*}{\coohoi{}} & Policy alone & 46.0 & 42.1 & -- & 44.1 & \textbf{45.9} & -- & 40.5 & 38.0 & -- & 37.1 & 43.1 & 48.5 & 40.9 & \textbf{40.0} & -- & 41.8 & 40.9 & 57.6 & 39.5 & 40.5 & 55.1 & 37.8 & \textbf{41.1} & \textbf{50.0} & 39.2 & 37.9 & \textbf{49.1} & 40.8 & 41.1 & 52.0 \\
 & \tdmpc{} & 50.5 & 49.5 & -- & 48.7 & 46.9 & -- & 47.9 & 47.4 & -- & 37.1 & \textbf{38.4} & 51.9 & 46.2 & 47.1 & -- & 48.4 & 45.5 & \textbf{49.0} & 45.4 & 48.1 & -- & 43.4 & 44.7 & 56.8 & 45.6 & 46.4 & -- & 45.9 & 46.0 & 52.5 \\
 & \seqwm{} & 49.7 & 48.9 & -- & 47.4 & 47.6 & -- & 46.5 & 42.3 & -- & 37.2 & 39.6 & 55.2 & 44.4 & 46.6 & -- & 43.6 & \textbf{39.5} & -- & 43.3 & 39.9 & 56.0 & 40.4 & 43.7 & -- & 41.8 & 41.6 & 55.7 & 43.8 & 43.3 & 55.6 \\
 & \textbf{\carlwm{} (Ours)} & \textbf{43.2} & \textbf{34.7} & -- & \textbf{42.2} & 46.7 & -- & \textbf{35.2} & \textbf{36.9} & 54.2 & \textbf{35.4} & 41.3 & \textbf{39.5} & \textbf{37.2} & \textbf{40.0} & 52.5 & \textbf{38.1} & \textbf{39.5} & 51.1 & \textbf{35.1} & \textbf{38.4} & \textbf{50.7} & \textbf{34.4} & 41.2 & \textbf{50.0} & \textbf{34.5} & \textbf{36.5} & 52.4 & \textbf{37.2} & \textbf{39.5} & \textbf{50.1} \\
\bottomrule
\end{tabular}}
\end{table*}

\subsection{Qualitative comparison}
\label{sec:supp-qualitative}
\begin{figure*}
\centering
\includegraphics[width=0.7\textwidth]{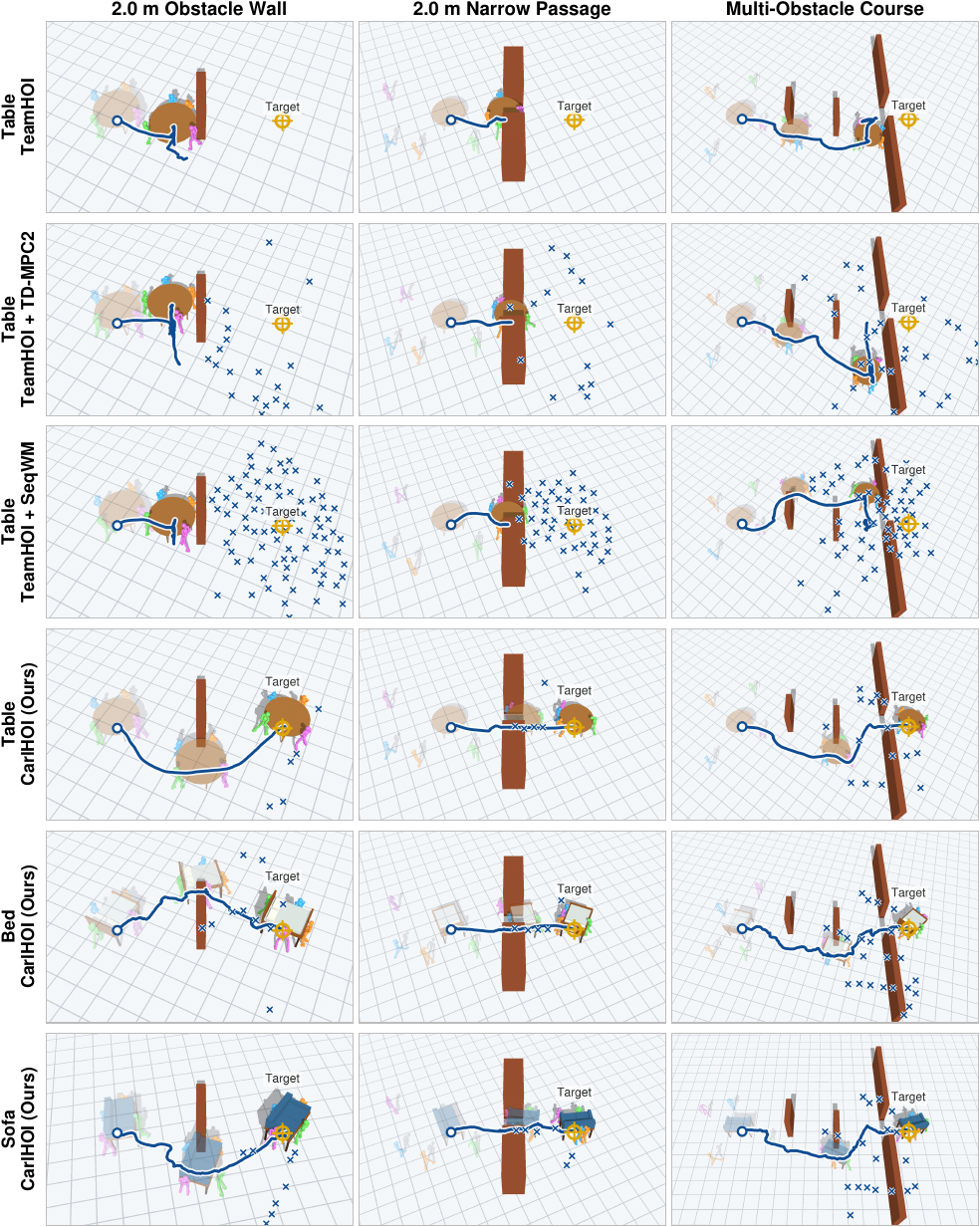}
\caption{\textbf{Qualitative results across tasks with four humanoids.} The first four rows compare the same table-transport episodes. The last two rows show CarlHOI transporting the unseen bed and sofa. Earlier poses are faded; blue lines trace the object center, and crosses mark temporary targets.}
\label{fig:supp-composites}
\end{figure*}

\Cref{fig:supp-composites} shows the same episode of each task under the four planners with \teamhoi{}. The
policy alone stalls at the wall and the gate. \tdmpc{}~\cite{hansen2024tdmpc2} and \seqwm{}~\cite{zhao2026seqwm} issue aim points that change direction between
decisions (and, for \seqwm{}, differ between humanoids), and the team does not reach the target. After pickup, \carlwm{} issues a
small number of consistent temporary targets that lead the team around the wall, through the gate and along the
course.

\begin{figure*}
\centering
\includegraphics[width=0.8\textwidth]{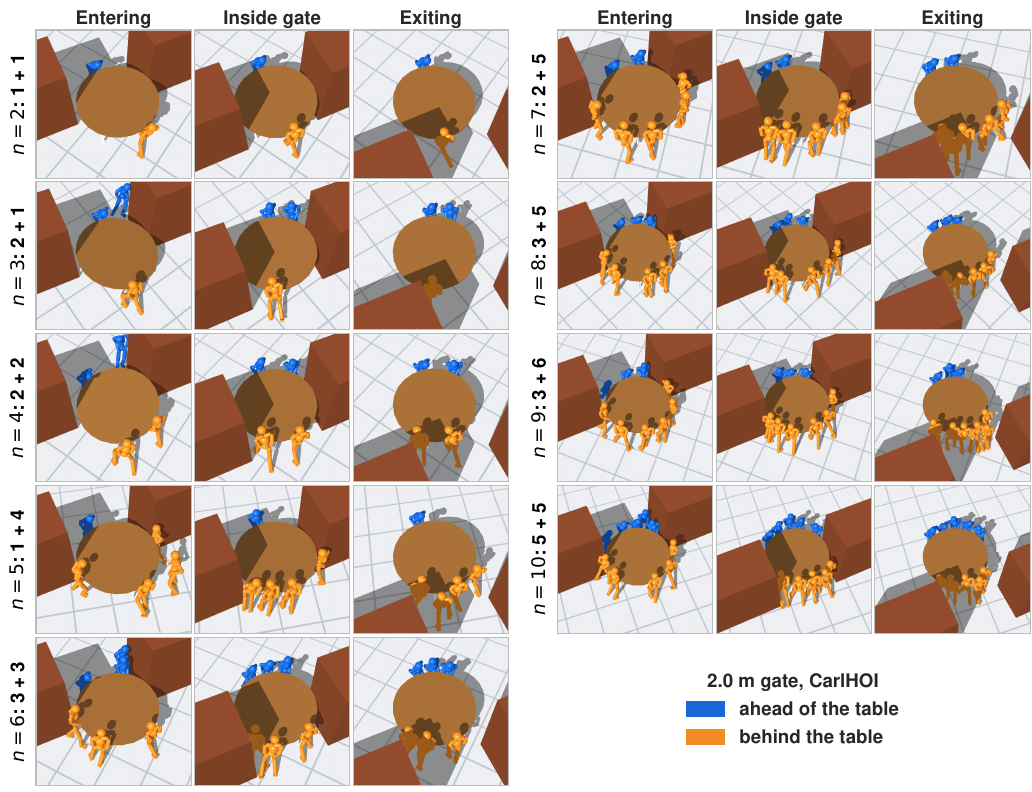}
\caption{\textbf{Team reformation for $n=2,\dots,10$} (\carlwm{} steering \teamhoift{}, table, 2.0\,m gate). One row
per team size; the row label gives the number of humanoids in front of and behind the table inside the gate. Blue:
ahead of the table center; orange: behind.}
\label{fig:supp-reformation}
\end{figure*}

\subsection{Team reformation across team sizes}
\label{sec:supp-reformation}
\Cref{fig:supp-reformation} extends the main-text reformation figure to every team size from 2 to 10, with one
successful crossing of the 2.0\,m gate per team size; this gate leaves no clearance for the table. At every size, the
team splits into a front and a rear group along the route and keeps holding the table through the gate. The split
differs between team sizes, from one humanoid on each end with $n=2$ to five and five with $n=10$, and is often
uneven (two in front and one behind with $n=3$, one and four with $n=5$, three and six with $n=9$). These are
selected episodes and do not show how often each split occurs.

\subsection{Other metrics}
\label{sec:supp-metrics}
\Cref{tab:supp-metrics} reports the rates of support loss and non-arrival, together with the final object-to-target distance. With \teamhoift{}, most of \carlwm{}'s gain comes from episodes that would
otherwise not arrive within the time limit: on the Obstacle Wall, these drop from 40.4\% to 2.8\%, and on the course
from 61.0\% to 32.1\%, while the final distance to the target shrinks from 2.25 to 0.55\,m and from 4.41 to 2.37\,m.
With TeamHOI-FT, loss of support accounts for approximately 30–36\% of episodes in the Narrow Passage across all planners. The baseline planners mainly
increase the share of episodes that do not arrive.

\section{Implementation}
\begin{itemize}[leftmargin=*,itemsep=1pt,topsep=2pt]
  \item Checkpoints: all models and evaluations use the same simulator settings and frozen skill checkpoints; the
  checkpoint hash is recorded with every run.
  \item Hardware: all training and evaluation run on 8 NVIDIA H100 GPUs and 1 NVIDIA A100 GPU.
  \item No method uses hand-written rules, privileged information or outcome-based tuning on test conditions.
\end{itemize}

\section{Benchmark and evaluation protocol}
\label{sec:supp-bench}

\paragraph{Task.}
\begin{itemize}[leftmargin=*,itemsep=1pt,topsep=2pt]
  \item Setup: $n$ simulated humanoids pick up an object and carry it to a target 7.5\,m away (13\,m on the multi-obstacle course task),
  in Isaac Gym at 30\,Hz.
  \item Objects: the round table used for training (diameter 2.0\,m, height 0.85\,m, 78\,kg), and the unseen bed ($1.8\times1.4$\,m)
  and sofa ($1.8\times1.2$\,m), both 50\,kg.
  \item Start: the humanoids start dispersed around the object and must first approach and lift it.
  \item Success: the object center comes within 0.03\,m of the target (\emph{arrival}).
  \item End: arrival, a failure (a humanoid falls, the object is dropped, or the simulator's own termination fires), or
  the time limit: 30\,s for the table (52\,s on the course task) and 60\,s for the bed and sofa.
  \item Physics, termination and the success rule are those of the released \teamhoi{} task~\cite{lionar2026teamhoi}, unchanged everywhere.
\end{itemize}

\paragraph{Obstacle tasks.} 
\begin{itemize}[leftmargin=*,itemsep=1pt,topsep=2pt]
  \item Narrow Passage: two 3.0\,m-wide, 1.2\,m-deep pieces with a gap of 2.1 or 2.0\,m, leaving 10\,cm or no clearance for
  the 2.0\,m table.
  \item Obstacle Wall: one wall 0.4\,m deep and 2.2\,m high, centered on the route, with a span of 1.0--3.0\,m. The team must
  carry the table around it.
  \item Multi-Obstacle Course: a 13\,m route with a 2.0\,m wall at 3.8\,m shifted 0.5\,m to one side, an identical wall at 7.3\,m
  shifted 0.5\,m to the other side, and a 2.00\,m gap at 11.0\,m between two 6.0\,m-wide, 0.4\,m-deep pieces; a mirrored layout balances the sides. The time limit scales with the route to 52\,s.
\end{itemize}

\paragraph{Planner training conditions.} The 43 collection runs, all with four humanoids and \teamhoift{}:
\begin{itemize}[leftmargin=*,itemsep=1pt,topsep=2pt]
  \item gates of 2.1\,m;
  \item walls of span 1.25, 1.5 and 1.75\,m, each at lateral offsets $-0.3$, $0$ and $+0.3$\,m.
\end{itemize}
Skill execution is deterministic in some collection runs and stochastic in others; this is recorded per run.

\begin{table*}[t]
\centering
\caption{\textbf{Carrying the unseen sofa across tasks and team sizes.} Layout and aggregation as in \cref{tab:main-table-object}: W: Obstacle Wall (\teamhoi{} and \coohoi{}: 1.0 and 2.0--3.0\,m), P: Narrow Passage (2.1 and 2.0\,m gates), M: Multi-Obstacle Course (two layouts). $^\dagger$Planners are trained with $n=4$, \teamhoift{} and the table only. Time: mean arrival time of successful episodes (limit 60\,s); --: no successful episode.}
\label{tab:supp-sofa}
\setlength{\tabcolsep}{1.3pt}
\resizebox{\textwidth}{!}{%
\begin{tabular}{@{}ll@{\hspace{5pt}}ccc@{\hspace{5pt}}ccc@{\hspace{5pt}}ccc@{\hspace{5pt}}ccc@{\hspace{5pt}}ccc@{\hspace{5pt}}ccc@{\hspace{5pt}}ccc@{\hspace{5pt}}ccc@{\hspace{5pt}}ccc@{\hspace{5pt}}ccc@{}}
\toprule
 & & \multicolumn{27}{c}{Team size $n$} & \\
\cmidrule(lr){3-29}
 & & \multicolumn{3}{c}{2} & \multicolumn{3}{c}{3} & \multicolumn{3}{c}{4$^\dagger$} & \multicolumn{3}{c}{5} & \multicolumn{3}{c}{6} & \multicolumn{3}{c}{7} & \multicolumn{3}{c}{8} & \multicolumn{3}{c}{9} & \multicolumn{3}{c}{10} & \multicolumn{3}{c}{Mean} \\
\cmidrule(lr){3-5}\cmidrule(lr){6-8}\cmidrule(lr){9-11}\cmidrule(lr){12-14}\cmidrule(lr){15-17}\cmidrule(lr){18-20}\cmidrule(lr){21-23}\cmidrule(lr){24-26}\cmidrule(lr){27-29}\cmidrule(lr){30-32}
Policy & Planner & W & P & M & W & P & M & W & P & M & W & P & M & W & P & M & W & P & M & W & P & M & W & P & M & W & P & M & W & P & M \\
\midrule
\multicolumn{32}{@{}l}{\textit{Success rate (\%) $\uparrow$}} \\
\midrule
\multirow{4}{*}{\teamhoi{}} & Policy alone & 5.4 & 2.1 & 0.0 & 27.3 & 21.5 & 0.4 & 35.1 & 35.4 & 2.9 & 42.2 & 32.8 & 5.1 & 39.7 & 30.7 & 7.0 & 39.3 & 21.7 & 7.4 & 35.4 & 16.0 & 3.1 & 38.1 & 16.0 & 4.3 & 37.3 & 13.9 & \textbf{4.3} & 33.3 & 21.1 & 3.8 \\
 & \tdmpc{} & 2.4 & 1.6 & 0.0 & 23.1 & 16.4 & 0.0 & 32.6 & 25.4 & 0.0 & 36.1 & 18.9 & 0.8 & 31.4 & 15.0 & 0.0 & 23.8 & 13.3 & 1.2 & 19.3 & 8.0 & 0.4 & 15.0 & 5.9 & 0.8 & 16.3 & 5.9 & 0.4 & 22.3 & 12.3 & 0.4 \\
 & \seqwm{} & 3.0 & 0.2 & 0.0 & 25.1 & 8.8 & 0.0 & 19.1 & 23.1 & 0.2 & 25.4 & 17.8 & 1.6 & 21.0 & 13.1 & 1.2 & 18.6 & 9.0 & 1.2 & 15.5 & 8.4 & 0.4 & 15.6 & 6.4 & 0.4 & 13.8 & 6.4 & 0.8 & 17.5 & 10.4 & 0.6 \\
 & \textbf{\carlwm{} (Ours)} & \textbf{10.3} & \textbf{4.5} & 0.0 & \textbf{44.2} & \textbf{27.5} & \textbf{1.6} & \textbf{60.6} & \textbf{41.4} & \textbf{11.3} & \textbf{59.2} & \textbf{43.4} & \textbf{10.2} & \textbf{57.8} & \textbf{35.0} & \textbf{12.9} & \textbf{49.5} & \textbf{27.9} & \textbf{7.8} & \textbf{43.8} & \textbf{17.2} & \textbf{5.9} & \textbf{43.4} & \textbf{17.0} & \textbf{5.9} & \textbf{42.0} & \textbf{14.3} & 3.9 & \textbf{45.6} & \textbf{25.3} & \textbf{6.6} \\
\cmidrule(l){2-32}
\multirow{4}{*}{\teamhoift{}$^\dagger$} & Policy alone & 14.5 & 18.6 & \textbf{0.4} & 46.0 & 39.3 & 2.7 & 54.1 & 51.6 & 12.3 & 63.0 & 54.5 & 15.6 & 63.0 & 50.8 & 14.1 & 68.8 & 47.9 & 8.6 & 66.9 & 41.4 & 5.1 & 66.4 & 39.5 & 3.9 & \textbf{67.6} & 38.7 & 3.5 & 56.7 & 42.4 & 7.4 \\
 & \tdmpc{} & 9.4 & 13.1 & 0.0 & 43.4 & 29.3 & 0.4 & 47.3 & 39.1 & 1.2 & 48.7 & 38.9 & 0.4 & 47.7 & 36.9 & 0.8 & 45.1 & 31.6 & 1.2 & 38.8 & 25.8 & 0.4 & 35.7 & 24.2 & 0.8 & 36.9 & 22.1 & 0.0 & 39.2 & 29.0 & 0.6 \\
 & \seqwm{} & 11.0 & 2.0 & 0.0 & 43.1 & 18.2 & 0.4 & 31.9 & 35.4 & 1.6 & 38.6 & 29.7 & 3.1 & 36.5 & 30.3 & 2.3 & 36.6 & 27.3 & 1.2 & 32.5 & 18.9 & 1.6 & 32.0 & 15.4 & 0.4 & 32.7 & 16.4 & 1.2 & 32.8 & 21.5 & 1.3 \\
 & \textbf{\carlwm{} (Ours)} & \textbf{33.3} & \textbf{26.2} & 0.0 & \textbf{60.8} & \textbf{49.2} & \textbf{5.5} & \textbf{72.0} & \textbf{55.7} & \textbf{20.9} & \textbf{73.4} & \textbf{60.2} & \textbf{24.2} & \textbf{73.8} & \textbf{59.2} & \textbf{20.3} & \textbf{72.6} & \textbf{53.3} & \textbf{16.8} & \textbf{69.1} & \textbf{55.5} & \textbf{7.0} & \textbf{66.6} & \textbf{52.7} & \textbf{7.4} & 62.8 & \textbf{41.0} & \textbf{5.5} & \textbf{64.9} & \textbf{50.3} & \textbf{12.0} \\
\cmidrule(l){2-32}
\multirow{4}{*}{\coohoi{}} & Policy alone & 0.2 & 0.0 & 0.0 & 6.5 & 5.5 & 0.0 & 3.9 & 5.5 & 0.0 & 19.6 & 15.2 & 0.8 & 21.0 & 13.9 & 0.0 & 15.5 & 9.4 & 0.4 & 28.8 & 10.9 & 0.0 & 43.0 & 21.7 & 1.2 & 32.2 & 19.7 & 1.2 & 19.0 & 11.3 & 0.4 \\
 & \tdmpc{} & 0.0 & 0.8 & 0.0 & 3.5 & 2.9 & 0.0 & 1.7 & 1.6 & 0.0 & 9.5 & 6.1 & 0.0 & 12.4 & 6.8 & 0.0 & 8.8 & 7.4 & 0.4 & 14.2 & 6.8 & 0.0 & 26.9 & 14.3 & 0.0 & 22.4 & 13.1 & 0.0 & 11.0 & 6.6 & 0.0 \\
 & \seqwm{} & 0.1 & 0.0 & 0.0 & 5.1 & 2.1 & 0.0 & 1.0 & 3.1 & 0.0 & 10.4 & 7.4 & 0.0 & 12.2 & 7.6 & 0.0 & 9.5 & 7.2 & 0.4 & 11.8 & 6.4 & 0.0 & 26.8 & 16.6 & 0.4 & 21.2 & 16.0 & 0.4 & 10.9 & 7.4 & 0.1 \\
 & \textbf{\carlwm{} (Ours)} & \textbf{1.3} & \textbf{1.6} & 0.0 & \textbf{10.1} & \textbf{7.6} & 0.0 & \textbf{7.0} & \textbf{10.6} & 0.0 & \textbf{25.2} & \textbf{22.9} & \textbf{1.6} & \textbf{33.3} & \textbf{19.7} & \textbf{1.2} & \textbf{20.8} & \textbf{15.2} & 0.4 & \textbf{38.2} & \textbf{23.0} & \textbf{2.7} & \textbf{56.3} & \textbf{32.0} & \textbf{3.5} & \textbf{56.5} & \textbf{35.0} & \textbf{2.3} & \textbf{27.6} & \textbf{18.6} & \textbf{1.3} \\
\midrule
\multicolumn{32}{@{}l}{\textit{Time to arrival (s), successful episodes only $\downarrow$}} \\
\midrule
\multirow{4}{*}{\teamhoi{}} & Policy alone & 52.2 & 44.9 & -- & 45.8 & \textbf{43.7} & 55.4 & 41.9 & 36.4 & \textbf{52.3} & 39.5 & 36.0 & 50.7 & 38.9 & 35.7 & 48.3 & 37.8 & 35.2 & 53.4 & 39.3 & 38.5 & 51.5 & 37.9 & 35.4 & \textbf{44.7} & 37.5 & \textbf{33.4} & \textbf{47.5} & 41.2 & 37.7 & 50.5 \\
 & \tdmpc{} & 53.5 & 51.9 & -- & 50.7 & 48.7 & -- & 48.5 & 46.4 & -- & 46.3 & 48.3 & 57.2 & 47.7 & 43.7 & -- & 45.9 & 45.5 & 55.9 & 46.2 & 47.5 & 54.0 & 45.8 & 40.4 & 56.1 & 44.7 & 44.4 & 56.3 & 47.7 & 46.3 & 55.9 \\
 & \seqwm{} & 52.5 & \textbf{44.1} & -- & 47.3 & 47.2 & -- & 46.7 & 41.3 & 59.4 & 41.7 & 40.4 & 51.5 & 42.4 & 38.9 & \textbf{37.0} & 41.9 & 41.0 & \textbf{48.0} & 40.0 & 38.8 & 58.7 & 41.2 & 39.9 & 56.0 & 40.6 & 40.4 & 49.2 & 43.8 & 41.3 & 51.4 \\
 & \textbf{\carlwm{} (Ours)} & \textbf{48.9} & 48.2 & -- & \textbf{42.6} & 45.2 & \textbf{53.8} & \textbf{38.0} & \textbf{35.0} & 53.9 & \textbf{36.2} & \textbf{35.4} & \textbf{50.6} & \textbf{35.7} & \textbf{34.6} & 50.6 & \textbf{36.8} & \textbf{34.4} & 50.9 & \textbf{36.5} & \textbf{35.4} & \textbf{47.1} & \textbf{37.7} & \textbf{31.9} & 46.0 & \textbf{35.6} & 38.0 & 47.7 & \textbf{38.6} & \textbf{37.6} & \textbf{50.1} \\
\cmidrule(l){2-32}
\multirow{4}{*}{\teamhoift{}$^\dagger$} & Policy alone & 49.8 & \textbf{45.4} & 43.0 & 41.7 & \textbf{39.7} & 52.2 & 37.9 & 34.4 & 51.5 & 34.7 & 34.2 & 46.8 & 35.9 & 33.8 & 49.7 & 36.2 & 34.9 & 50.0 & 37.5 & 36.0 & 54.5 & 37.6 & 34.8 & 51.5 & 37.7 & 35.9 & 49.5 & 38.8 & 36.6 & 49.9 \\
 & \tdmpc{} & 51.9 & 52.0 & -- & 48.5 & 48.6 & \textbf{50.1} & 46.9 & 45.0 & 53.2 & 45.6 & 45.9 & 59.5 & 45.4 & 46.8 & 54.4 & 46.6 & 46.7 & 51.8 & 47.9 & 47.9 & 53.8 & 48.0 & 45.6 & 51.4 & 47.9 & 45.9 & -- & 47.6 & 47.2 & 53.5 \\
 & \seqwm{} & 50.3 & 50.9 & -- & 43.7 & 46.8 & 51.8 & 43.4 & 40.2 & 53.6 & 40.3 & 40.0 & 51.5 & 40.7 & 38.1 & 52.9 & 41.7 & 37.8 & \textbf{42.4} & 42.1 & 38.7 & 54.1 & 41.6 & 39.0 & \textbf{40.6} & 42.0 & 37.5 & \textbf{44.8} & 42.9 & 41.0 & \textbf{49.0} \\
 & \textbf{\carlwm{} (Ours)} & \textbf{46.8} & 45.6 & -- & \textbf{38.5} & 41.3 & 54.2 & \textbf{32.9} & \textbf{32.2} & \textbf{50.1} & \textbf{31.3} & \textbf{31.1} & \textbf{46.1} & \textbf{31.5} & \textbf{30.6} & \textbf{48.7} & \textbf{33.1} & \textbf{32.5} & 48.1 & \textbf{35.1} & \textbf{35.5} & \textbf{51.9} & \textbf{34.8} & \textbf{34.3} & 54.0 & \textbf{34.8} & \textbf{34.5} & 51.7 & \textbf{35.4} & \textbf{35.3} & 50.6 \\
\cmidrule(l){2-32}
\multirow{4}{*}{\coohoi{}} & Policy alone & \textbf{36.4} & -- & -- & \textbf{40.7} & 41.3 & -- & 46.3 & 42.4 & -- & 42.5 & 39.8 & 54.9 & 43.9 & 43.4 & -- & 40.8 & 42.1 & \textbf{41.0} & 43.6 & 41.8 & -- & 40.2 & 39.5 & 55.8 & 40.8 & 40.9 & 51.8 & 41.7 & 41.4 & 50.9 \\
 & \tdmpc{} & -- & \textbf{48.0} & -- & 47.1 & 49.5 & -- & 49.8 & 52.3 & -- & 44.4 & 45.8 & -- & 45.7 & 46.1 & -- & 47.1 & 45.7 & 53.1 & 48.8 & 48.3 & -- & 46.5 & 44.7 & -- & 46.2 & 48.1 & -- & 47.0 & 47.6 & 53.1 \\
 & \seqwm{} & 50.4 & -- & -- & 43.6 & 43.7 & -- & 48.9 & 47.1 & -- & 43.4 & \textbf{38.2} & -- & 46.3 & 43.7 & -- & 43.0 & 42.1 & 48.9 & 44.1 & 41.9 & -- & 42.8 & 40.5 & \textbf{51.8} & 43.2 & 39.7 & \textbf{45.8} & 45.1 & 42.1 & \textbf{48.9} \\
 & \textbf{\carlwm{} (Ours)} & 45.2 & 50.6 & -- & 43.0 & \textbf{39.3} & -- & \textbf{44.9} & \textbf{40.9} & -- & \textbf{40.2} & 39.7 & \textbf{46.3} & \textbf{40.9} & \textbf{40.3} & 55.0 & \textbf{39.8} & \textbf{38.2} & 43.9 & \textbf{39.8} & \textbf{37.9} & 54.4 & \textbf{36.0} & \textbf{36.1} & 52.4 & \textbf{36.6} & \textbf{34.9} & 52.2 & \textbf{40.7} & \textbf{39.8} & 50.7 \\
\bottomrule
\end{tabular}}
\end{table*}

\begin{table}[t]
\centering
\caption{\textbf{Success rate (\%) per obstacle, table}, averaged over $n=2,\dots,10$. Walls by span (m), gates by gap (m), course layouts A and B (mirrored). Bold: best planner per policy and column. Training conditions of \carlwm{}: 1.25--1.75\,m walls, 2.1\,m gate, $n=4$, \teamhoift{}, table.}
\label{tab:supp-subtask-table}
\setlength{\tabcolsep}{2pt}
\resizebox{\columnwidth}{!}{%
\begin{tabular}{@{}llcccc@{\hspace{4pt}}cc@{\hspace{4pt}}cc@{}}
\toprule
 & & \multicolumn{4}{c}{Obstacle Wall} & \multicolumn{2}{c}{Narrow Passage} & \multicolumn{2}{c}{Course} \\
\cmidrule(lr){3-6}\cmidrule(lr){7-8}\cmidrule(l){9-10}
Policy & Planner & 1.0 & 2.0 & 2.5 & 3.0 & 2.1 & 2.0 & A & B \\
\midrule
\multirow{4}{*}{\teamhoi{}} & Policy alone & 79.4 & 48.0 & 30.3 & 19.7 & 42.5 & 17.3 & 13.8 & 16.8 \\
 & \tdmpc{} & 43.8 & 29.0 & 20.6 & 15.6 & 10.1 & 1.7 & 1.0 & 1.9 \\
 & \seqwm{} & 49.0 & 28.6 & 21.2 & 13.6 & 21.6 & 5.5 & 8.3 & 6.3 \\
 & \textbf{\carlwm{} (Ours)} & \textbf{84.7} & \textbf{80.8} & \textbf{78.9} & \textbf{76.9} & \textbf{47.3} & \textbf{20.0} & \textbf{26.6} & \textbf{29.0} \\
\cmidrule(l){2-10}
\multirow{4}{*}{\teamhoift{}$^\dagger$} & Policy alone & 79.6 & 44.0 & 26.7 & 16.0 & 59.5 & 29.6 & 23.9 & 20.2 \\
 & \tdmpc{} & 41.1 & 28.3 & 19.9 & 15.4 & 13.7 & 1.8 & 0.9 & 1.6 \\
 & \seqwm{} & 46.2 & 29.9 & 18.1 & 11.4 & 31.4 & 8.5 & 9.9 & 9.2 \\
 & \textbf{\carlwm{} (Ours)} & \textbf{93.1} & \textbf{87.8} & \textbf{87.1} & \textbf{84.5} & \textbf{71.5} & \textbf{39.2} & \textbf{46.7} & \textbf{48.3} \\
\cmidrule(l){2-10}
\multirow{4}{*}{\coohoi{}} & Policy alone & 67.8 & 45.5 & 33.8 & 24.0 & 17.7 & 7.6 & 5.8 & 5.8 \\
 & \tdmpc{} & 34.2 & 21.1 & 15.1 & 11.3 & 3.8 & 0.7 & 0.6 & 0.3 \\
 & \seqwm{} & 36.9 & 19.8 & 13.7 & 8.7 & 7.7 & 2.3 & 3.5 & 3.0 \\
 & \textbf{\carlwm{} (Ours)} & \textbf{83.4} & \textbf{81.4} & \textbf{75.0} & \textbf{65.4} & \textbf{29.2} & \textbf{15.3} & \textbf{16.8} & \textbf{15.6} \\
\bottomrule
\end{tabular}}
\end{table}

\begin{table}[t]
\centering
\caption{\textbf{Success rate (\%) per obstacle, bed}, averaged over $n=2,\dots,10$. Walls by span (m), gates by gap (m), course layouts A and B (mirrored). Bold: best planner per policy and column. Training conditions of \carlwm{}: 1.25--1.75\,m walls, 2.1\,m gate, $n=4$, \teamhoift{}; the bed is unseen.}
\label{tab:supp-subtask-bed}
\setlength{\tabcolsep}{2pt}
\resizebox{\columnwidth}{!}{%
\begin{tabular}{@{}llcccc@{\hspace{4pt}}cc@{\hspace{4pt}}cc@{}}
\toprule
 & & \multicolumn{4}{c}{Obstacle Wall} & \multicolumn{2}{c}{Narrow Passage} & \multicolumn{2}{c}{Course} \\
\cmidrule(lr){3-6}\cmidrule(lr){7-8}\cmidrule(l){9-10}
Policy & Planner & 1.0 & 2.0 & 2.5 & 3.0 & 2.1 & 2.0 & A & B \\
\midrule
\multirow{4}{*}{\teamhoi{}} & Policy alone & 64.4 & 49.2 & 39.7 & 31.6 & 26.7 & 20.7 & 4.7 & 4.5 \\
 & \tdmpc{} & 45.3 & 35.1 & 28.9 & 23.9 & 16.5 & 13.5 & 0.3 & 0.6 \\
 & \seqwm{} & 45.0 & 28.8 & 20.4 & 14.9 & 14.5 & 11.3 & 1.3 & 0.8 \\
 & \textbf{\carlwm{} (Ours)} & \textbf{69.4} & \textbf{63.0} & \textbf{57.1} & \textbf{49.0} & \textbf{32.1} & \textbf{24.6} & \textbf{10.4} & \textbf{8.6} \\
\cmidrule(l){2-10}
\multirow{4}{*}{\teamhoift{}$^\dagger$} & Policy alone & 81.7 & 68.6 & 55.4 & 43.6 & 52.6 & 42.9 & 9.2 & 9.2 \\
 & \tdmpc{} & 66.9 & 55.2 & 43.4 & 34.9 & 38.9 & 28.4 & 0.6 & 0.5 \\
 & \seqwm{} & 63.2 & 40.8 & 31.6 & 23.6 & 29.6 & 23.4 & 1.7 & 1.3 \\
 & \textbf{\carlwm{} (Ours)} & \textbf{84.6} & \textbf{77.9} & \textbf{69.4} & \textbf{64.1} & \textbf{64.2} & \textbf{51.9} & \textbf{14.1} & \textbf{13.4} \\
\cmidrule(l){2-10}
\multirow{4}{*}{\coohoi{}} & Policy alone & 47.0 & 36.8 & 28.0 & 22.2 & 20.7 & 16.8 & 0.6 & 0.8 \\
 & \tdmpc{} & 31.0 & 25.8 & 21.7 & 16.1 & 12.3 & 11.3 & 0.2 & 0.2 \\
 & \seqwm{} & 35.7 & 21.4 & 17.0 & 13.2 & 13.4 & 10.5 & 0.1 & 0.3 \\
 & \textbf{\carlwm{} (Ours)} & \textbf{54.5} & \textbf{51.8} & \textbf{50.3} & \textbf{45.8} & \textbf{31.4} & \textbf{25.6} & \textbf{4.2} & \textbf{2.6} \\
\bottomrule
\end{tabular}}
\end{table}

\begin{table}[t]
\centering
\caption{\textbf{Success rate (\%) per obstacle, sofa}, averaged over $n=2,\dots,10$. Walls by span (m), gates by gap (m), course layouts A and B (mirrored). Bold: best planner per policy and column. Training conditions of \carlwm{}: 1.25--1.75\,m walls, 2.1\,m gate, $n=4$, \teamhoift{}; the sofa is unseen.}
\label{tab:supp-subtask-sofa}
\setlength{\tabcolsep}{2pt}
\resizebox{\columnwidth}{!}{%
\begin{tabular}{@{}llcccc@{\hspace{4pt}}cc@{\hspace{4pt}}cc@{}}
\toprule
 & & \multicolumn{4}{c}{Obstacle Wall} & \multicolumn{2}{c}{Narrow Passage} & \multicolumn{2}{c}{Course} \\
\cmidrule(lr){3-6}\cmidrule(lr){7-8}\cmidrule(l){9-10}
Policy & Planner & 1.0 & 2.0 & 2.5 & 3.0 & 2.1 & 2.0 & A & B \\
\midrule
\multirow{4}{*}{\teamhoi{}} & Policy alone & 48.7 & 35.3 & 27.9 & 21.4 & 22.5 & 19.7 & 3.9 & 3.8 \\
 & \tdmpc{} & 32.6 & 23.4 & 18.8 & 14.3 & 14.1 & 10.4 & 0.3 & 0.5 \\
 & \seqwm{} & 30.1 & 18.6 & 12.6 & 8.6 & 11.0 & 9.7 & 0.7 & 0.5 \\
 & \textbf{\carlwm{} (Ours)} & \textbf{55.7} & \textbf{47.6} & \textbf{42.2} & \textbf{37.0} & \textbf{26.9} & \textbf{23.7} & \textbf{7.7} & \textbf{5.5} \\
\cmidrule(l){2-10}
\multirow{4}{*}{\teamhoift{}$^\dagger$} & Policy alone & 68.0 & 57.0 & 45.1 & 36.9 & 44.6 & 40.3 & 7.1 & 7.6 \\
 & \tdmpc{} & 48.4 & 37.7 & 28.8 & 24.4 & 30.6 & 27.4 & 0.4 & 0.7 \\
 & \seqwm{} & 46.4 & 30.2 & 21.2 & 16.4 & 22.8 & 20.2 & 1.6 & 1.0 \\
 & \textbf{\carlwm{} (Ours)} & \textbf{72.0} & \textbf{66.3} & \textbf{57.3} & \textbf{51.2} & \textbf{53.6} & \textbf{47.0} & \textbf{12.7} & \textbf{11.2} \\
\cmidrule(l){2-10}
\multirow{4}{*}{\coohoi{}} & Policy alone & 28.0 & 19.1 & 16.4 & 12.5 & 11.6 & 11.0 & 0.4 & 0.3 \\
 & \tdmpc{} & 15.4 & 11.5 & 9.2 & 8.0 & 7.6 & 5.7 & 0.0 & 0.1 \\
 & \seqwm{} & 17.8 & 11.2 & 8.7 & 5.9 & 8.3 & 6.5 & 0.1 & 0.2 \\
 & \textbf{\carlwm{} (Ours)} & \textbf{34.2} & \textbf{27.9} & \textbf{26.1} & \textbf{22.3} & \textbf{18.6} & \textbf{18.6} & \textbf{1.1} & \textbf{1.5} \\
\bottomrule
\end{tabular}}
\end{table}

\paragraph{Evaluation (Testing) conditions.} Everything outside these ranges is for testing:
\begin{itemize}[leftmargin=*,itemsep=1pt,topsep=2pt]
  \item the 2.0\,m gate;
  \item the 1.0, 2.0, 2.5 and 3.0\,m walls;
  \item the Multi-Obstacle Course;
  \item every team size other than four;
  \item the bed and the sofa;
  \item every transport policy other than \teamhoift{} (\teamhoi{} and \coohoi{}~\cite{gao2024coohoi}).
\end{itemize}
No training data is drawn from any test condition.

\paragraph{Team-size benchmark.}
\begin{itemize}[leftmargin=*,itemsep=1pt,topsep=2pt]
  \item Cells: gates of 2.1 and 2.0\,m and walls of 1.0, 1.25, 1.5, 1.75, 2.0, 2.5 and 3.0\,m, each with
  $n=2,\dots,10$.
  \item Objects: the bed and the sofa use the same scenes and initial states as the table.
  \item Pairing: an episode is identified by its scene seed and environment index, which fix the geometry and the
  initial state. All methods therefore run on identical episodes.
  \item Execution: policies sample actions from their action distribution, as during training.
\end{itemize}

\paragraph{Metrics and statistics.}
\begin{itemize}[leftmargin=*,itemsep=1pt,topsep=2pt]
  \item Success rate and time to arrival of successful episodes, as in the main text. The Mean columns report the unweighted average across team sizes.
  \item Outcome breakdown and final distance to the target (\cref{sec:supp-metrics}).
  \item Statistics: paired differences with 95\% confidence intervals from 10{,}000 bootstrap resamples of episodes,
  stratified by scene seed. 
\end{itemize}

\section{Planners}
\paragraph{Shared interface.}
\begin{itemize}[leftmargin=*,itemsep=1pt,topsep=2pt]
  \item Aim point: every 0.5\,s a planner may replace the target that the skill observes with
  $\mathbf{a} = \mathbf{p} + s\,R(\theta)(\mathbf{g}-\mathbf{p})$. Here $\mathbf{p}$ is the object position,
  $\mathbf{g}$ the true target, $|\theta|\le 45^\circ$ and $s\in[0.5,1.5]$. The aim distance is clipped to
  0.5--9\,m.
  \item Unchanged: the skill, its reward, the success rule and the true target.
  \item Inputs, the same for all planners: recent object and humanoid states, the target, the obstacle pieces as
  boxes, and the episode time.
\end{itemize}

\paragraph{\carlwm{}.}
\begin{itemize}[leftmargin=*,itemsep=1pt,topsep=2pt]
  \item Commands: nine aim commands --- the unchanged target, rotations of $\pm15^\circ$, $\pm30^\circ$ and
  $\pm45^\circ$, and scales of 0.5 and 1.5.
  \item Planning: each command is rolled out in latent space for 1.5\,s (nine steps of 1/6\,s) and scored by
  \cref{eq:score}; the highest score is executed. The reference bonus $\omega=0.02$ favours unchanged guidance when
  scores are close. The planner engages once the object has been picked up.
  \item Observation: the last six states at 1/6\,s intervals. For the object: position, height, heading, velocity and
  tilt. For each humanoid, relative to the object: position, height, heading, velocity, hand-to-rim distance. Also the previous commands, the target in the route frame, the obstacle pieces as boxes relative to the
  object, and the episode time.
  \item Architecture: a 3-layer transformer ($d=128$) over tokens for the object, each humanoid (shared weights), each
  obstacle piece and the context. Absent humanoids and pieces are masked. The output is pooled into a 128-dimensional
  SimNorm latent.
  \item Anchoring: the dynamics and the progress, failure, contact and arrival heads are anchored on the unchanged
  command, $z' = \mathrm{SimNorm}\big(D(z) + C(z,a) - C(z,a_0)\big)$, and likewise for the heads. The terminal value
  is an ensemble of five heads. In total, 1{,}524{,}749 parameters.
  \item Data: 43 collection runs on the training conditions, giving 703{,}118 samples. At each branch point the
  simulator state is saved, and each command of the run's library is rolled out for 1.5\,s from it.
  \begin{itemize}[leftmargin=*,itemsep=1pt,topsep=1pt]
    \item Library: five commands in 18 runs (the first runs, at the 2.1\,m gate) and nine commands in 25 runs.
    \item Branching: every 3\,s in 41 runs and every 1.5\,s in 2 runs.
    \item Behaviour: the unchanged command with probability 0.4 or a random command; later, an $\epsilon$-greedy
    ($\epsilon=0.3$) behaviour of the previous planner.
  \end{itemize}
  \item Losses: multi-step latent consistency (weight 20), progress (1), events (1), value (0.5, Monte-Carlo returns),
  and a sibling ranking loss on the score differences of commands simulated from the same state (1). Steps are
  weighted by $\rho=0.9$.
  \item Optimiser: AdamW, learning rate $3\times10^{-4}$ with the encoder at $0.3\times$, weight decay $10^{-4}$,
  batch 256, input noise 0.05. At most 40 epochs; the checkpoint with the best validation ranking loss is kept, with
  patience 10.
\end{itemize}

\paragraph{Baseline principles.}
\begin{itemize}[leftmargin=*,itemsep=1pt,topsep=2pt]
  \item Shared: the aim-point interface, the training conditions used in the 43 collection runs, the simulator budget that produced \carlwm{}'s data
  (21{,}364{,}224 environment steps, branch rollouts included), and a model size matched within 0.4\%.
  \item Training: online, from scratch, with each method's own exploration, on the reward that \carlwm{}'s score
  encodes. Per decision: progress toward the true target, $-3$ for a failure, $-0.3/9$ per wall-contact step, in
  units of 5\,cm. Arrival and failure are terminal; the time limit is not.
  \item Timing: decisions every 0.5\,s from the start of the episode.
  \item Evaluation: the final agent is evaluated, without exploration noise or checkpoint selection.
\end{itemize}

\paragraph{\tdmpc{}~\cite{hansen2024tdmpc2}.}
\begin{itemize}[leftmargin=*,itemsep=1pt,topsep=2pt]
  \item Implementation: the official implementation, ported with identical update and planning math to run inside our
  simulator process. The episodic variant, with a termination head.
  \item Action: one continuous command $(\theta,s)$ for the team.
  \item Network: the 1M preset (latent 128, encoder width 256, two Q-functions), with MLP width 375 to match the
  parameter count (1{,}522{,}234).
  \item Planner: MPPI with 512 samples, 6 iterations, 64 elites, 24 policy-prior trajectories, horizon 3 decisions
  (1.5\,s) and temperature 0.5. The mean is warm-started from the previous plan.
  \item Training: discount 0.95 (the official heuristic); 1{,}000 seed transitions with uniform actions; then one
  update per collected transition, batch 256, Adam with learning rate $3\times10^{-4}$ and the encoder at $0.3\times$.
  \item Observation: 687 numbers --- the last six team--object states (humanoid slots padded to ten), the target, the
  two obstacle pieces, the episode time and the previous action.
\end{itemize}

\paragraph{\seqwm{}.~\cite{zhao2026seqwm}}
\begin{itemize}[leftmargin=*,itemsep=1pt,topsep=2pt]
  \item Settings: the paper's hyperparameter tables~\cite{zhao2026seqwm} where they give a value; otherwise the official code
  and configuration.
  \item Agents: each humanoid is an agent with its own command $(\theta,s)$, which steers only that humanoid's goal
  input. Agents plan in sequence, each conditioning on its predecessors' plans.
  \item Model per agent: a SimNorm encoder; dynamics and reward conditioned on the agent's own action and its
  predecessors' actions; two critics; a tanh-Gaussian actor.
  \item Team sizes: one set of weights is shared by all humanoids. Messages carry the actions of the three
  immediately preceding humanoids, with presence bits. The critic is per agent, as in the paper's formulation, so one
  model covers $n=2,\dots,10$. Latent width 320; 1{,}529{,}523 parameters.
  \item Planner: horizon 3, 6 iterations, 512 samples including 24 actor trajectories, 64 elites, temperature 0.5.
  \item Training: batch 1{,}000, 20-step returns, replay buffer $10^6$, learning rate $5\times10^{-4}$ with the
  encoder at $0.3\times$, discount 0.96. A warm-up of 50{,}000 random transitions and 2{,}500 updates, then one
  update per four transitions, the ratio of the official configuration.
  \item Observation: each humanoid sees the \tdmpc{} observation with its own humanoid listed first.
\end{itemize}

\begin{table}
\centering
\caption{\textbf{Planners compared.} All are trained with four humanoids on the same conditions with \teamhoift{}
and act through the same aim-point interface. \emph{Branched}: counterfactual rollouts of several commands from restored
simulator states. $\pi$: learned policy prior.}
\footnotesize
\setlength{\tabcolsep}{2.5pt}
\begin{tabular}{@{}llll@{}}
\toprule
 & \carlwm{} & \tdmpc{} & \seqwm{} \\
\midrule
Parameters & 1.525M & 1.522M & 1.530M \\
Training & offline & online & online \\
Data & branched & own rollouts & own rollouts \\
Sim.\ steps & 21.36M & 21.36M & 21.36M \\
Action & 9 commands & team $(\theta,s)$ & per-agent $(\theta,s)$ \\
Planner & exhaustive & MPPI + $\pi$ & sequential MPPI \\
Lookahead & 1.5\,s + $V$ & 1.5\,s + $Q$ & 1.5\,s + $Q$ \\
Acts & once carried & from the start & from the start \\
Team size & widened & padded & shared agents \\
\bottomrule
\end{tabular}
\label{tab:supp-planners}
\end{table}

\section{\carlwm{} training details}
\label{sec:supp-carlwm}
\begin{itemize}[leftmargin=*,itemsep=1pt,topsep=2pt]
  \item Optimizer: AdamW. The encoder uses a $0.3\times$ learning rate; with the full learning rate the transformer
  encoder did not train stably. Training runs for at most 40 epochs. The model is selected by the ranking
  loss~\eqref{eq:rank} on the five-command validation branch points, with early stopping after 10 epochs without
  improvement.
  \item Team size: the model is trained with four humanoid slots. After training, its humanoid token capacity is
  expanded to ten; no humanoid parameters are added, since all humanoid tokens share weights. Slots of absent humanoids
  are masked out of the attention, and active humanoids are never masked, so the output for four humanoids is
  unchanged and teams of 2--10 use the same weights. The expanded model also has a team-size input to the context token
  whose weights are initialized to zero, so it has no effect.
  \item Ranking-loss branch points: the first collection rounds branched the five commands $u^{0}$, $\pm15^\circ$,
  $\times0.5$ and $\times1.5$; later rounds branched all nine. \Cref{eq:rank} is computed on two sets, all branch points
  restricted to these five commands and the nine-command branch points (16{,}535 in training) over all nine, each
  sampled at 64 branch points per minibatch; the two terms are added with equal weight.
  \item Event loss: the binary cross-entropy of each event uses a positive-class weight equal to its ratio of
  negative to positive training steps, clipped to $[1,50]$, since failures, contacts and arrivals are rare.
  \item Target of $u^{0}$: the unchanged command is simulated twice at each branch point, and its realized return in
  \cref{eq:rank} is the mean of the two.
\end{itemize}

\section{Evaluated combinations}
\begin{itemize}[leftmargin=*,itemsep=1pt,topsep=2pt]
  \item Every transport policy (\teamhoi{}, \teamhoift{}, \coohoi{}) alone and with \carlwm{}, \tdmpc{} and
  \seqwm{}, on all tasks, team sizes and objects, on identical episodes.
  \item \carlwm{} is never retrained or tuned for another policy, team size, obstacle or object.
\end{itemize}

\subsection{Ablation protocol}
\label{sec:supp-ablation-protocol}
\begin{itemize}[leftmargin=*,itemsep=1pt,topsep=2pt]
  \item Components removed, one at a time:
  \begin{itemize}[leftmargin=*,itemsep=1pt,topsep=1pt]
    \item the long-horizon value (kept in training, dropped from the planning score);
    \item the ranking loss;
    \item the relational encoder, replaced by a parameter-matched MLP over the same per-entity inputs (width 514;
    1{,}523{,}963 parameters);
    \item counterfactual anchoring;
  \end{itemize}
  \item Training: each ablated model is retrained with \carlwm{}'s data, recipe and seed 0.
  \item Evaluation: four humanoids and \teamhoift{}. Narrow Passage: the 2.1\,m and 2.0\,m gates, pooled with equal weight, for the table, bed and sofa. Obstacle Wall (table): all seven spans
  of 1.0--3.0\,m. Multi-Obstacle Course (table): the mirrored layout.
  \item The ablation scenes for the gates, the 1.0 and 2.0\,m walls and the course are separate from those of
  \cref{tab:main-table-object}, so the full model's values differ slightly from that table.
\end{itemize}

\section{Low-level policies}
\begin{itemize}[leftmargin=*,itemsep=1pt,topsep=2pt]
  \item \textbf{\teamhoi{}}: the released multi-agent transformer policy, trained on the open carrying task with
  adversarial motion priors. It has never seen gates or walls.
  \item \textbf{\teamhoift{}}: initialised from \teamhoi{} (policy, value function and discriminator) and fine-tuned
  with PPO for 4{,}096{,}000 transitions on the 2.1\,m gate with four humanoids. The task reward is replaced by
  target-arrival shaping, and a geometry encoder of the passage is added as a zero-initialised residual. Settings:
  128 environments, horizon 32, minibatch 512, 6 epochs, learning rate $4\times10^{-6}$, $\gamma=0.995$.
  \item \textbf{\coohoi{}}: re-implemented following the specification in~\cite{lionar2026teamhoi}.
  \begin{itemize}[leftmargin=*,itemsep=1pt,topsep=1pt]
    \item Network: \teamhoi{}'s transformer with cross-attention replaced by self-attention and no teammate tokens.
    \item Contact points: fixed designated points, evenly spaced on the rim, with the designated point listed first in
    the observation.
    \item Reward: \teamhoi{}'s, with the formation terms replaced by the approach-angle reward.
    \item Stage 1: a single agent with halved table mass and frictionless table legs, for 3 paper-days
    ($2.00\times10^9$ transitions).
    \item Stage 2: \coohoi-4, for 6 paper-days ($1.32\times10^9$ transitions), validated four times per paper-day. It
    stops at the paper's four-agent success of 94.5\%.
    \item Team sizes: one model, trained with four agents, is used for every team size, as for \teamhoi{}.
    \item Multi-GPU training, when used, is data-parallel with the totals of environments and minibatch samples
    unchanged, so each update equals the single-GPU update.
    \item Acceptance: on \teamhoi{}'s open-task protocol (10{,}240 episodes per team size), after calibrating the
    evaluator on the released \teamhoi{} checkpoint.
  \end{itemize}
\end{itemize}

\end{document}